\documentclass{article}

\usepackage[preprint]{neurips_2025}

\usepackage[utf8]{inputenc}
\usepackage[T1]{fontenc}
\usepackage{hyperref}
\usepackage{url}
\usepackage{booktabs}
\usepackage{amsfonts,amsmath,amssymb}
\usepackage{nicefrac}
\usepackage{microtype}
\usepackage[table]{xcolor}
\usepackage{graphicx}
\usepackage{tikz}
\usetikzlibrary{arrows.meta,positioning,fit,backgrounds,calc,shapes.geometric,patterns,decorations.pathreplacing}
\usepackage[numbers,square,sort&compress]{natbib}
\usepackage{caption}
\usepackage{pgfplots}
\usepackage{float}
\usepackage{placeins}
\usepackage{tabularx,array}
\usepackage{multirow}      
\definecolor{oursblue}{RGB}{232,240,254}
\hypersetup{
 colorlinks=true,
 linkcolor=red,
 citecolor=cyan,
 filecolor=magenta,
 urlcolor=magenta,
}
\makeatletter
\renewcommand{\@noticestring}{Technical Report.}
\makeatother

\usepackage{fancyhdr}
\renewcommand{\headwidth}{\textwidth}

\renewcommand{\headrulewidth}{0.5pt}
\renewcommand{\headrule}{\vspace{2pt}\hbox to\headwidth{\color{black}\leaders\hrule height \headrulewidth\hfill}}
\pgfplotsset{compat=1.18}

\AtBeginDocument{\raggedbottom\brokenpenalty=10000\relax}

\title{
  TuringPAI: A Foundation for On-Device Physical AI
}

\author{
\colorbox{white}{\includegraphics[height=0.8em]{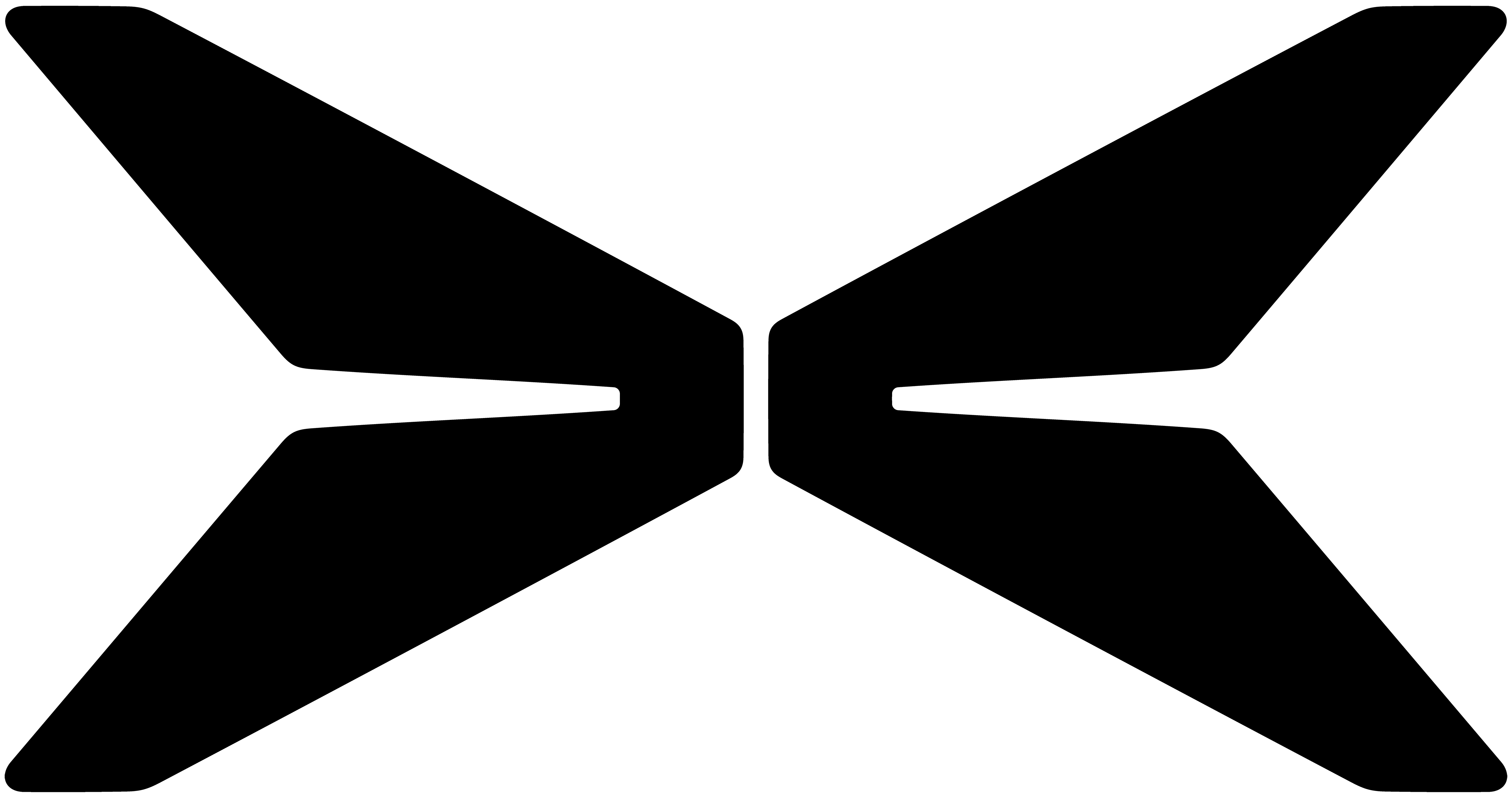}} Foundation Model Team, XPeng Inc.\\
}
\date{Technical report draft · October 2026}

\begin{document}
\maketitle

\begin{abstract}

Physical AI requires models that understand spatial relationships and dynamic environments and translate observations into actions under limited onboard compute. We present TuringPAI, a family of models for on-device Physical AI comprising TuringPAI-VL for shared vision-language understanding, TuringPAI-Drive for autonomous driving, and TuringPAI-Embodied for robotic manipulation. TuringPAI-VL is a 0.65B model built on HybridViT-MLQ, which combines convolutional local aggregation with Transformer-based global modeling. Multi-scale Latent Queries integrate spatial features and observation history within the same visual Transformer, providing a compact, structured interface to the language model. Training combines general vision-language data with geometric, cross-view, and temporal supervision. TuringPAI-VL supports general multimodal understanding and improves quantitative geometric perception over Qwen2-VL-2B, while providing a shared initialization for the two downstream VLA models. For autonomous driving, TuringPAI-Drive maintains all evaluated ADE/FDE metrics within 1.2\% of a full-token HybridViT baseline using the MLQ interface and achieves comparable aggregate scores in an in-house simulation evaluation. Using XPeng Native Runtime (XNRT) on a single NVIDIA Orin-X SoC, TuringPAI-Drive generates trajectories at 12\,Hz in on-vehicle operation. For robotic manipulation, TuringPAI-Embodied achieves a 96.8\% average success rate across the four LIBERO suites. These results demonstrate a shared vision-language foundation for driving and robotic manipulation under practical on-device constraints.
\end{abstract}

\begin{center}
    \vspace{-0.6em}
    \includegraphics[width=\textwidth]{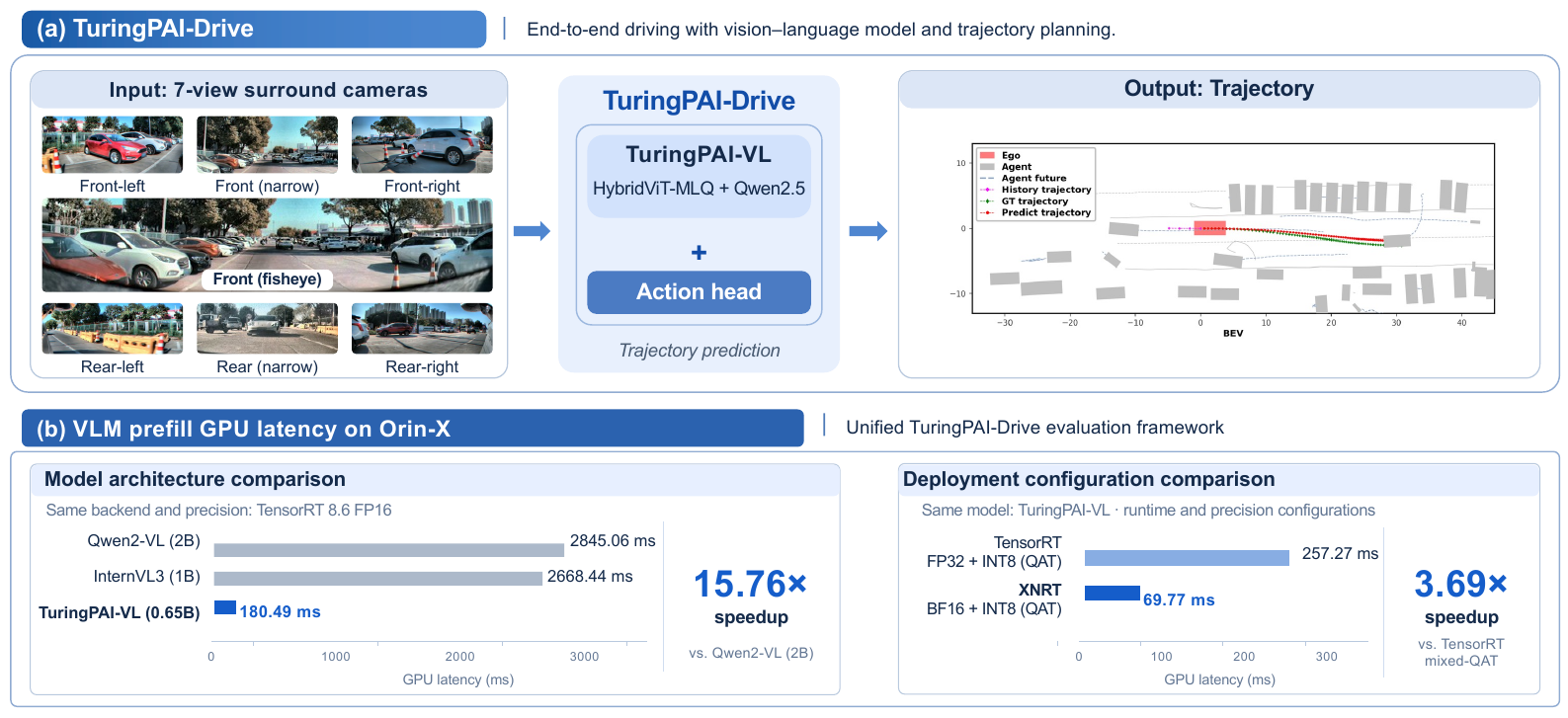}
    \captionof{figure}
    {\textbf{TuringPAI-Drive and on-device VLM inference.}
(a) Seven-camera trajectory prediction with TuringPAI-VL and an action head.
(b) GPU-side VLM prefill latency on one NVIDIA Orin-X SoC in the TuringPAI-Drive evaluation framework.
\emph{Left:} TensorRT~8.6 FP16 timings with model-specific inputs (not accuracy-matched).
\emph{Right:} On the same TuringPAI-VL backbone, XNRT BF16+INT8 is $3.69\times$ faster than TensorRT FP32+INT8 (both with QAT).
}

    \label{fig:vla_overview}
    \vspace{-0.8em}
\end{center}

\clearpage
\section{Introduction}
\label{sec:intro}

Physical AI systems must understand changing environments and translate observations into decisions and actions under real-world computational and latency constraints~\cite{shukor2025smolvla, black2026real, zhang2026turingllm}. Vision-language models (VLMs) provide a reusable foundation by connecting visual perception with language instructions and reasoning~\cite{liu2023visual, wang2024qwen2vl, bai2025qwen3vl}. Autonomous driving, a mature and practically important application of Physical AI, serves as a major setting for developing and evaluating such models~\cite{team2026xcot, tian2024drivevlm}. It requires spatial understanding, reasoning over observations across cameras and time, and timely responses within limited onboard resources. Robotic manipulation presents complementary demands on object geometry, interaction state, and instruction following~\cite{black2024pi0}.

Visual inputs in Physical AI applications often consist of high-resolution observations collected across multiple cameras and time steps~\cite{caesar2020nuscenes, khazatsky2024droid}.
These observations contain substantial redundancy, yet local geometry and small temporal changes can affect downstream actions.
VLM pipelines that encode each observation as a dense patch sequence can spend substantial computation on spatially correlated content~\cite{bolya2023tome, shang2025llava}.
Forwarding separate visual sequences for successive frames further expands the language-model context with overlapping scene information~\cite{huang2025prunevid, shen2026fastvid}.
Under limited onboard compute and strict latency constraints, this redundancy increases the cost of using rich observation histories.
These requirements motivate VLM designs that aggregate local evidence before global modeling and integrate complementary information across spatial scales and time, while retaining the cues needed for downstream action.

Architecture alone, however, does not ensure that a VLM learns the physical information required by these tasks.
General image--text alignment and instruction learning support broad semantic understanding, but do not necessarily provide explicit supervision for metric geometry or temporal changes.
For example, a caption may identify a vehicle ahead without specifying its distance or whether the gap is closing, although both quantities matter for driving decisions.
Prior work shows that explicit spatial and temporal training tasks can strengthen the corresponding capabilities of VLMs~\cite{chen2024spatialvlm, ren2024timechat}.
Architecture and training data thus play complementary roles:
the architecture organizes observations efficiently, while the data guide the learning of information relevant to physical interaction.

To address these challenges, we present TuringPAI, a family of models for on-device Physical AI built around a shared vision-language foundation.
At its core, TuringPAI-VL combines HybridViT-MLQ with a training mixture tailored to Physical AI.
TuringPAI-VL provides the pretrained foundation for TuringPAI-Drive and TuringPAI-Embodied, which adapt it to autonomous driving and robotic manipulation, respectively.

HybridViT-MLQ organizes local features, scene context, and observation history into a multi-scale representation of the current observation.
HybridViT first uses a convolutional front end to form locally condensed patch tokens, over which the visual Transformer models global relationships.
We then introduce Multi-scale Latent Queries (MLQ), a hierarchy of feature-conditioned queries that aggregates visual evidence within the same Transformer.
The queries occupy spatial grids at multiple resolutions and share a spatiotemporal positional encoding with patch tokens.
For video, one query hierarchy is initialized from the current frame and aggregates evidence from both current and preceding frames.
Only the updated query features are passed to the language model.
For a fixed query layout, this interface incorporates observation history without increasing the output sequence length.
HybridViT thus reduces the sequence length used for visual encoding, while MLQ organizes multi-scale and temporal evidence within a controlled language-model input budget.
MLQ can also be adapted to other ViT backbones with compatible feature, positional, and attention interfaces.
As an additional deployment benefit, the convolutional front end supports heterogeneous execution on NVIDIA Orin-X hardware~\cite{nvidiaDriveAGXOrin}, including DLA and GPU overlap in our implementation.

To train TuringPAI-VL, we complement general vision--language data with geometric, cross-view, and temporal tasks.
Following the data-construction approach of VLM3~\cite{cai2026vlm3}, we use pixel-depth questions and camera-conditioned 3D object localization to supervise distance, position, size, and orientation, providing geometric cues for driving clearance and object placement. Cross-view correspondence and relative camera-pose data support associations across observations, while multi-frame driving and manipulation examples provide signals about motion and state changes.

We evaluate TuringPAI-VL on general multimodal benchmarks and spatial
and geometric understanding tasks relevant to Physical AI.
Driving evaluations of TuringPAI-Drive cover open-loop trajectory
prediction and XPeng's in-house simulation.
For deployment, we use XPeng Native Runtime (XNRT), our in-house inference
runtime that combines mixed-precision execution with hardware-aware
optimization, enabling 12~Hz trajectory generation on a single NVIDIA Orin-X SoC.
TuringPAI-Drive is scheduled for rollout to XPeng vehicles equipped with NVIDIA Orin-X as part of the VLA~2.0 system.
Beyond driving, TuringPAI-Embodied, initialized from TuringPAI-VL,
achieves a 96.8\% average success rate across the four LIBERO~\cite{liu2023libero} suites.
These results support TuringPAI-VL as a shared vision-language
foundation for driving and manipulation under practical deployment
constraints.

\section{Methodology}
\label{sec:arch}

\begin{figure}[H]
  \centering
  \includegraphics[width=\textwidth]{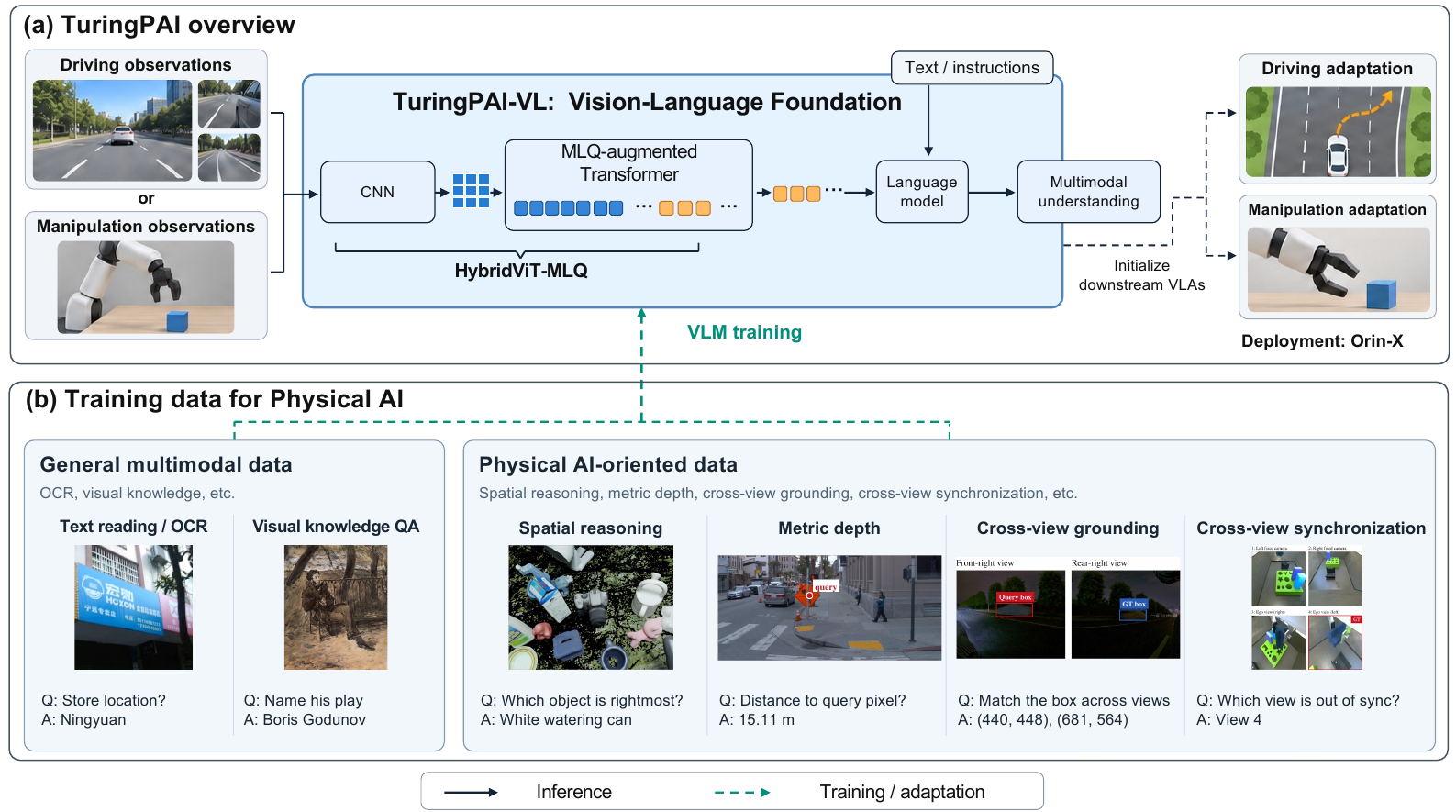}
  \caption{
  \textbf{Overview of TuringPAI-VL.}
  (a) HybridViT-MLQ combines convolutional local aggregation with
  a visual Transformer augmented by latent queries, supplying
  multi-scale visual features to the language model.
  (b) General multimodal data are complemented by Physical-AI-oriented
  tasks, illustrated through spatial reasoning, metric depth,
  cross-view grounding, and cross-view synchronization examples.
  Solid arrows denote inference flow; dashed arrows denote training
  or downstream adaptation.
  The visual-to-language projector is omitted for clarity.
  }
  \label{fig:orinvlm-main-architecture}
\end{figure}

\begin{figure}[t] %
  \centering
  \includegraphics[width=\textwidth]{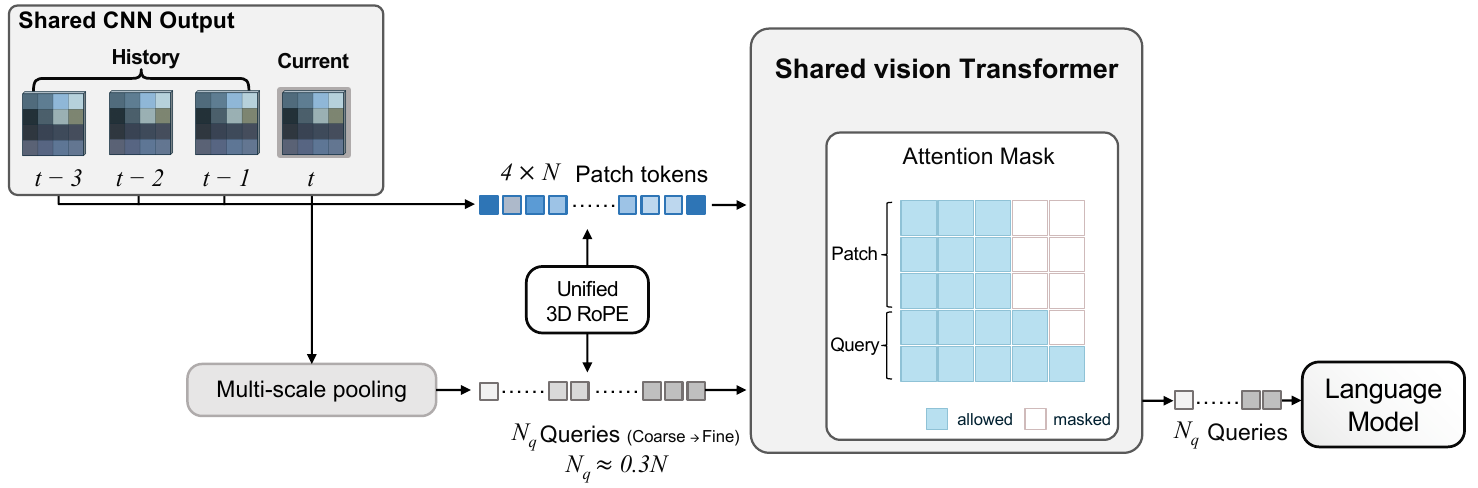}
  \caption{\textbf{Multi-scale Latent Queries (MLQ).} Current-frame features initialize coarse-to-fine queries, processed with observation-window patches using unified 3D RoPE. Masking preserves patch-to-patch attention while queries aggregate patch and earlier-query features; only updated queries pass to the language model. Rows and columns of the attention mask correspond to query and key positions, respectively. The linear projection used to initialize latent queries is omitted for clarity. Frame labels show temporal order; token counts are schematic.}
  \label{fig:mlq-architecture}
\end{figure}

TuringPAI-VL is the shared vision-language foundation of the
TuringPAI family.
Its design centers on three elements: HybridViT, Multi-scale Latent
Queries (MLQ), and training data for Physical AI.
First, HybridViT combines convolutional aggregation of local image
structure with Transformer-based global modeling over condensed patch
tokens (Section~\ref{sec:hybridvit}).
Second, MLQ aggregates multi-scale spatial features and observation
history into a structured visual interface for the language model
within the same visual Transformer (Section~\ref{sec:mlq}).
Third, we supplement general vision--language data with geometric,
cross-view, and temporal tasks to develop capabilities relevant to
physical interaction (Section~\ref{sec:training-data}).
Together, these elements support spatial and temporal understanding
while enabling efficient visual processing for downstream Physical AI
tasks.
Figure~\ref{fig:orinvlm-main-architecture} presents the overall
architecture, downstream adaptation, and representative training
examples.
Figure~\ref{fig:mlq-architecture} details the MLQ mechanism.

\subsection{HybridViT: Local Aggregation and Global Modeling}
\label{sec:hybridvit}

HybridViT combines convolutional local feature extraction with
Transformer-based global modeling.
The convolutional front end in Figure~\ref{fig:orinvlm-main-architecture}(a) aggregates local image content into locally condensed patch tokens.
The encoder supports dynamic-resolution inputs by adjusting image height and width to multiples of $128$.
The resized image is partitioned into $128\times128$ computation regions.
With a total spatial stride of $32$ along each axis, the convolutional front end produces a $4\times4$ feature grid for each region.
These regional features are reassembled into a full-image feature map $\mathbf F\in\mathbb R^{H_v\times W_v\times C}$,
where $H_v=\widetilde H/32$ and $W_v=\widetilde W/32$ for a resized input of dimensions $\widetilde H\times\widetilde W$.
We flatten this grid and project its features to the Transformer
embedding dimension:
\begin{equation}
\begin{aligned}
\mathbf V
&=
\phi_{\mathrm{in}}
\bigl(\operatorname{Flatten}(\mathbf F)\bigr)
\in\mathbb R^{N\times D},\\
N
&=
H_vW_v
=
\frac{\widetilde H\widetilde W}{32^2}.
\end{aligned}
\label{eq:hybridvit-tokenization}
\end{equation}
Here, $C$ and $D$ are the CNN and Transformer feature dimensions,
respectively.
The input projection $\phi_{\mathrm{in}}$ applies the same linear
map from $C$ to $D$ dimensions to each feature vector.
At the same input resolution, a patch-size-16 ViT~\cite{faghri2025mobileclip2, tschannen2025siglip} produces
$N_{16}=\widetilde H\widetilde W/16^2$ tokens.
HybridViT therefore supplies one quarter as many patch tokens to the Transformer.

The Transformer component follows the mixed-attention design of TuringViT~\cite{wu2026turingvit}, combining efficient and standard
self-attention.
It models relationships across the condensed patch sequence to
incorporate global scene context.
MLQ operates within this same Transformer, as described next.

\subsection{Multi-scale Latent Queries}
\label{sec:mlq}

\paragraph{Backbone reuse and query construction.}
Multi-scale Latent Queries (MLQ) provide a query-based plug-in that aggregates visual features within a pretrained Transformer.
MLQ reuses the backbone's attention and feed-forward weights without adding a separate Transformer stack.
It can be incorporated during VLM training without an additional visual pretraining stage.
The design is not specific to HybridViT; adaptation to another ViT requires compatible feature, positional, and attention interfaces.

In HybridViT-MLQ, multi-scale pooling and linear projection of
convolutional features initialize latent query grids at different
spatial resolutions, as illustrated in Figure~\ref{fig:mlq-architecture}.
The grid sizes adapt to the spatial dimensions of the input feature
map.
The grids are flattened and concatenated in coarse-to-fine order, forming a feature-conditioned query hierarchy spanning global, regional, and local spatial scales.

\paragraph{Unified spatiotemporal positioning.}
We organize video inputs into four-frame observation windows, taking the last frame in each window as the current observation.
Each window uses one latent-query hierarchy initialized from the current frame's features.
A single image is treated as a one-frame window, with queries initialized from its own features.
To position patch tokens and queries consistently across scales and frames, we extend the DINO-style 2D RoPE~\cite{simeoni2025dinov3} used by TuringViT to three dimensions by introducing a temporal coordinate.

\begin{samepage}
\noindent For a token at row $r$ and column $c$ of a spatial grid $g$, we assign the following coordinate for 3D RoPE:
\begin{equation}
\mathbf p_g(r,c;\tau)
=
\left(
\tau,\,
2\frac{r+\tfrac{1}{2}}{H_g}-1,\,
2\frac{c+\tfrac{1}{2}}{W_g}-1
\right),
\label{eq:mlq-shared-coordinates}
\end{equation}
where $H_g\times W_g$ is the grid resolution and $r,c$ are zero-based indices.
Cell centers in both patch and query grids are normalized to the same image domain $[-1,1]^2$, providing a common spatial reference across scales.
\par
\end{samepage}

Inspired by the use of physical time in
LLaVA-OneVision-2~\cite{an2026llavaonevision2}, the temporal coordinate $\tau$ records the actual acquisition timestamp in a consistent time unit.
Patch tokens retain the timestamps of their respective frames, while all latent queries use the current frame's timestamp.
This preserves elapsed time between observations and anchors the query hierarchy to the current scene.
In the single-image case, all tokens share one timestamp.

\paragraph{Query-based aggregation.}
The asymmetric attention mask in Figure~\ref{fig:mlq-architecture} prevents patch tokens from attending to latent queries, leaving the backbone's patch-to-patch attention pattern unchanged.
Latent queries attend to all input patch tokens and to themselves
and preceding queries in coarse-to-fine order.
This allows queries to aggregate visual evidence and coarser context
without feeding query states back into patch-token updates.
Only the updated query features are projected into the language embedding space, forming multi-scale visual tokens that represent the current observation with evidence from the available history.

Let $N$ denote the number of HybridViT patch tokens per frame and
$N_q$ the number of latent queries.
For $T$ frames with the same spatial resolution, the sequence
lengths are
\begin{equation}
\begin{aligned}
N_{\mathrm{ViT}} &= TN+N_q,\\
N_{\mathrm{LLM}}^{\mathrm{vis}} &= N_q,
\end{aligned}
\label{eq:mlq-token-budget}
\end{equation}
where $N_{\mathrm{LLM}}^{\mathrm{vis}}$ counts only the visual tokens
passed to the language model.
Our evaluated configurations use $T\in\{1,4\}$ and
$N_q\approx0.3N$.
For a fixed spatial resolution and query layout, the output token
count is independent of the number of frames in the observation
window.
For a single image, convolutional aggregation keeps the combined
patch-and-query sequence approximately one third as long as
patch-size-16 tokenization at the same input resolution.
The language model receives only $N_q$ visual tokens rather than
the full $TN$ patch sequence, consolidating multi-scale and
historical evidence within a controlled output budget.

\begin{figure}[!htbp]
  \centering
  \includegraphics[width=0.95\linewidth]{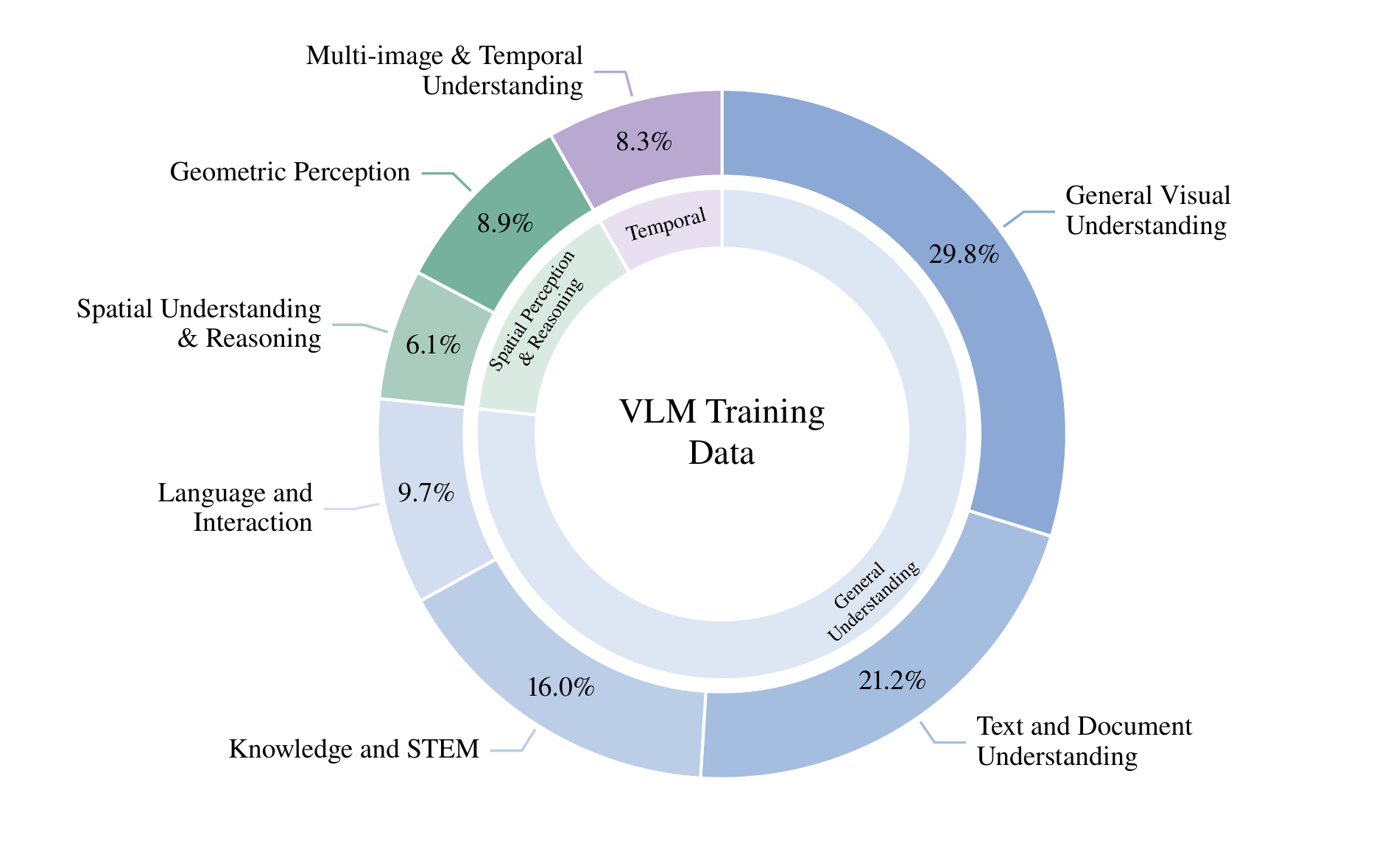}
  \caption{\textbf{Composition of the TuringPAI-VL training pool across general understanding, spatial perception and reasoning, and multi-image and temporal understanding.}}
  \label{fig:training-data-composition}
\end{figure}

\subsection{Training Data for Physical AI Capabilities}
\label{sec:training-data}

To train TuringPAI-VL as a foundation for downstream driving and
manipulation models, we combine general vision--language data
with tasks that provide explicit training signals for spatial
geometry, cross-view association, and temporal understanding.

The training data cover three complementary capabilities: general
understanding, spatial perception and reasoning, and multi-image and temporal
understanding. These capabilities are learned jointly within a single VLM.
The resulting model directly generates spatial predictions from visual inputs
and task prompts, without additional task-specific fine-tuning for spatial
evaluation.
Figure~\ref{fig:orinvlm-main-architecture}(b) illustrates representative training tasks.
Figure~\ref{fig:training-data-composition} summarizes the overall data-pool composition, and Appendix~\ref{app:spatial-supervision-cases} provides further examples.

\paragraph{General understanding.} General image--text and instruction data establish vision--language alignment and broad semantic understanding. They cover referring-expression grounding, visual question answering, text and document understanding, knowledge and STEM, and language interaction. These tasks support scene interpretation, environmental text reading, and instruction following, providing a semantic foundation for spatial, geometric, and temporal learning.

\paragraph{Spatial perception and reasoning.} Building on general visual grounding and spatial-relation understanding, we introduce quantitative geometric supervision following VLM3~\citep{cai2026vlm3}. Depth estimation and 3D object localization provide information about distance, size, and orientation, offering geometric cues for assessing following distances, safety margins, and driving clearance, as well as approaching and placing objects in embodied manipulation. Cross-view pixel correspondence constrains local matching of the same scene point, while camera-pose estimation supervises relative rotation and translation between observations, aiming to strengthen geometric association across cameras and moving viewpoints. These tasks encourage compact visual representations to capture both object semantics and fine-grained spatial structure.

\paragraph{Multi-image and temporal understanding.} We further incorporate multi-view question answering, videos, and multi-frame sequences from road-traffic and robotic-manipulation scenarios. Multi-view tasks emphasize object association and consistency across observations, training the model to combine complementary camera information and interpret scenes using both global layout and local details.
Temporal tasks focus on motion trends, state changes, and interaction dynamics, providing historical context for assessing approaching objects and relative motion in driving, as well as manipulation progress and task stages in robotics. This design complements MLQ's aggregation of historical information into current-observation representations, aiming to capture both spatial and dynamic information within a limited visual token budget.

\section{Experiments}
\label{sec:experiments}

This section evaluates three models within the TuringPAI family: TuringPAI-VL, the vision-language foundation model, and two domain-specific extensions, TuringPAI-Drive and TuringPAI-Embodied. TuringPAI-VL combines HybridViT-MLQ with a Qwen2.5-0.5B~\citep{yang2024qwen25} language model and is evaluated against comparable VLMs for general multimodal and physical-world understanding. TuringPAI-Drive extends the foundation model with a trajectory planning head. It maintains driving ADE/FDE metrics within 1.2\% of the full-token HybridViT baseline, with comparable driving-quality scores. On a single NVIDIA Orin-X SoC, TuringPAI-Drive generates trajectories at 12\,Hz in on-vehicle operation.  TuringPAI-Embodied integrates a flow-matching action expert with the same foundation model for robotic manipulation. It achieves 96.8\% average success on LIBERO, close to the published OpenPI $\pi_{0.5}$ result. On a single NVIDIA Orin-X SoC, its measured inference stages after CNN patch embedding sum to 168.3\,ms. Together, these results demonstrate the versatility of the TuringPAI foundation across autonomous driving and robotic manipulation. All on-device deployment and latency results reported in this work use a single NVIDIA Orin-X SoC.

\FloatBarrier

\subsection{TuringPAI-VL: Foundation Model Training and Evaluation}
\label{sec:exp-vlm}

TuringPAI-VL is the compact VLM designed to meet the on-device budget of XPeng's
Orin-X-based in-vehicle computing platform while supporting real-time inference. It pairs
HybridViT-MLQ (156M) with Qwen2.5-0.5B~\citep{yang2024qwen25}, for
approximately 0.65B parameters in total. Training proceeds in two stages:
visual-token alignment with the language model frozen, followed by joint
training on the full data pool. We then evaluate TuringPAI-VL on general multimodal
benchmarks, including image and video understanding, and on spatial-perception
benchmarks. 
The pretrained TuringPAI-VL checkpoint, including both the visual encoder and language model, initializes the driving and robotic-manipulation VLAs.

\subsubsection{Training Paradigm and Data Composition}
\label{sec:exp-vlm-setup}

To balance general-purpose vision-language capabilities with the
physical-world understanding required by Physical AI, TuringPAI-VL
adopts a two-stage training paradigm. The first stage establishes
vision-language alignment using a general image--text corpus, with
the language model frozen and only the visual encoder and adapter
trained. Building on this initialization, the second stage jointly
optimizes all model parameters using a broader image/video--text
mixture that combines general multimodal tasks with Physical-AI-oriented
supervision. In particular, the training data cover geometric and
spatial perception, cross-view understanding, and temporal reasoning,
alongside captioning, VQA, text and document understanding, knowledge,
and STEM. These tasks span both autonomous-driving environments and
robotic-manipulation scenarios. Furthermore, logit-level distillation
from a larger TuringViT-based VLM is incorporated into the second
stage to transfer its multimodal capabilities to the compact model.
Both stages use native dynamic resolution. Table~\ref{tab:vlm-setup}
summarizes the training configurations for both stages, including their
objectives, batch sizes, learning rates, and computational costs.

\begin{table}[htbp]
  \caption{\textbf{VLM training setup.} Stage~1 trains the encoder and adapter with the
  language model frozen; Stage~2 trains all parameters. $^{*}$The H800
  GPU-hours indicate total training cost, measured as the number of H800 GPUs
  multiplied by wall-clock training hours.}
  \label{tab:vlm-setup}
  \centering
  \footnotesize
  \setlength{\tabcolsep}{4pt}
  \renewcommand{\arraystretch}{1.12}
  \begin{tabular}{@{}lcc@{}}
    \toprule
    \textbf{Configuration} & \textbf{Stage 1 (alignment)} & \textbf{Stage 2 (joint)} \\
    \midrule
    Data type & general image--text & image/video--text: grounding, \\
    & & text-rich, spatial, temporal \\
    Resolution & native dynamic & native dynamic \\
    Objective & image--text alignment & joint + logit distillation \\
    Global batch & $2{,}048$ & $4{,}096$ \\
    Peak LR & $2.0{\times}10^{-4}$ & $5.0{\times}10^{-5}$ \\
    Final LR & $4.0{\times}10^{-5}$ & $5.0{\times}10^{-6}$ \\
    H800 GPU-hours$^{*}$ & $1{,}640$ & $9{,}870$ \\
    \bottomrule
  \end{tabular}
\end{table}

\subsubsection{Benchmark Results}
\label{sec:exp-vlm-results}

\renewcommand{\bottomfraction}{0.9}
\renewcommand{\topfraction}{0.9}
\begingroup

\begin{table}[t] %
  \caption{\textbf{VLM benchmark results for TuringPAI-VL and public models.} Bold and
  underlined mark the best and second-best among the three compared models.
  Results for Qwen2-VL-2B Base and InternVL3-1B Base are from our own evaluations.}
  \label{tab:vlm-1b}
  \centering
  \footnotesize
  \setlength{\tabcolsep}{3pt}
  \renewcommand{\arraystretch}{1.15}
  
  \begin{tabular}{llcc>{\columncolor{oursblue}}c}
    \toprule
    Group & Benchmark
      & \shortstack{Qwen2-VL-2B\\Base~\citep{wang2024qwen2vl}}
      & \shortstack{InternVL3-1B\\Base~\citep{zhu2025internvl3}}
      & TuringPAI-VL \\
    \midrule
    \multirow{2}{*}{\shortstack[l]{Grounding \\ \& Counting}} & RefCOCO avg~\citep{kazemzadeh2014referitgame, yu2016modeling} & 64.16 & \underline{77.88} & \textbf{80.91} \\
 & CountBenchQA~\citep{paiss2023teachingclip,beyer2024paligemma} & \textbf{83.57} & 56.67 & \underline{69.61} \\
    \cmidrule(lr){1-5}
    \multirow{7}{*}{\shortstack[l]{Text-rich \\ Perception}} & OCRBench V1~\citep{liu2024ocrbench} & \underline{724} & 716 & \textbf{727} \\
 & OCRBench V2 en~\citep{fu2025ocrbenchv2} & \textbf{32.59} & 25.29 & \underline{32.10} \\
 & OCRBench V2 cn~\citep{fu2025ocrbenchv2} & 12.79 & \underline{21.36} & \textbf{29.33} \\
 & ChartQA test~\citep{masry2022chartqa} & 54.52 & \underline{62.92} & \textbf{76.96} \\
 & TextVQA val~\citep{singh2019textvqa} & 64.55 & \textbf{70.15} & \underline{65.89} \\
 & DocVQA val~\citep{mathew2021docvqa} & \textbf{84.44} & 74.75 & \underline{80.52} \\
 & AI2D test~\citep{kembhavi2016diagram} & \textbf{72.12} & 63.60 & \underline{69.56} \\
    \cmidrule(lr){1-5}
    \multirow{5}{*}{\shortstack[l]{Multimodal \\ Reasoning}} & MMMU val~\citep{yue2024mmmu} & \textbf{38.89} & \underline{37.78} & 34.11 \\
 & MathVision mini~\citep{wang2024mathvision} & \underline{17.43} & 13.82 & \textbf{17.76} \\
 & MathVista mini~\citep{lu2024mathvista} & \underline{42.70} & 36.60 & \textbf{54.20} \\
 & DynaMath~\citep{zou2024dynamath} & 19.40 & \textbf{23.45} & \underline{19.56} \\
 & LogicVista~\citep{xiao2024logicvista} & 24.61 & \textbf{27.52} & \underline{26.40} \\
    \cmidrule(lr){1-5}
    \multirow{8}{*}{\shortstack[l]{General\\VQA}} & MMBench-EN~\citep{liu2023mmbench} & \textbf{69.93} & 63.83 & \underline{67.34} \\
 & MMBench-CN~\citep{liu2023mmbench} & \textbf{66.58} & 59.54 & \underline{65.33} \\
 & RealWorldQA~\citep{xai2024realworldqa} & \textbf{61.18} & 51.24 & \underline{57.91} \\
 & MMStar~\citep{chen2024mmstar} & \underline{46.40} & 42.67 & \textbf{47.40} \\
 & MM-Vet~\citep{yu2023mmvet} & \underline{39.17} & 32.80 & \textbf{39.91} \\
 & HallusionBench (aAcc)~\citep{guan2024hallusionbench} & 53.73 & \underline{54.36} & \textbf{54.99} \\
 & MME-RealWorld-Lite~\citep{zhang2024mmerealworld} & 28.71 & \textbf{38.56} & \underline{29.91} \\
 & SEEDBench\_IMG~\citep{li2023seedbench} & \textbf{72.26} & 68.14 & \underline{71.70} \\
    \cmidrule(lr){1-5}
    \multirow{3}{*}{\shortstack[l]{Video \\ Understanding}} & MVBench~\citep{li2024mvbench} & \underline{60.89} & 55.13 & \textbf{61.95} \\
 & MLVU~\citep{zhou2025mlvu} & \textbf{55.71} & 47.21 & \underline{51.50} \\
 & LongVideoBench val total~\citep{wu2024longvideobench} & \underline{45.47} & 40.24 & \textbf{45.60} \\
    \bottomrule
  \end{tabular}%
\end{table}
\endgroup

Following the three capability areas introduced in
Section~\ref{sec:training-data}, we evaluate general understanding, spatial
perception and reasoning, and multi-image and temporal understanding. We use
the same jointly trained 0.65B-parameter TuringPAI-VL checkpoint for all
tasks, without task-specific fine-tuning.

\paragraph{General understanding.} TuringPAI-VL supports a range of
image understanding tasks (Table~\ref{tab:vlm-1b}). We use InternVL3-1B Base~\citep{zhu2025internvl3} as the primary reference
because it is closer in parameter count to TuringPAI-VL.
TuringPAI-VL scores higher on chart interpretation
(76.96 vs. 62.92 on ChartQA), document understanding
(80.52 vs. 74.75 on DocVQA), and visual mathematical reasoning
(54.20 vs. 36.60 on MathVista mini). InternVL3-1B Base remains
stronger on TextVQA and MMMU. We include the larger Qwen2-VL-2B Base~\citep{wang2024qwen2vl}
as an additional reference.
TuringPAI-VL retains advantages in chart interpretation and visual
mathematical reasoning, while Qwen2-VL-2B Base is stronger in counting,
document understanding, and several general VQA tasks, as illustrated
by CountBenchQA, DocVQA, and MMBench. For 2D referring-expression grounding,
TuringPAI-VL achieves an average score of 80.91 on RefCOCO, outperforming
InternVL3-1B Base (77.88).

\paragraph{Spatial perception and reasoning.} The clearest advantages are in
depth and relative camera pose (Table~\ref{tab:spatial-perception}). On the
NYU Depth V2 test split~\citep{silberman2012indoor}, TuringPAI-VL achieves a
mean per-image absolute relative depth error (AbsRel) of 0.146, compared with
0.38 for InternVL3-1B Base. For cross-view point correspondence on ETH3D
image pairs~\citep{schops2017eth3d}, the gain is smaller: mean endpoint error
is 193.49 pixels, versus 199.05 for the same baseline. For relative pose, we
evaluate 180 image pairs from six scenes in 7-Scenes~\citep{shotton2013scene}
using the AUC@30$^\circ$ metric from Depth Anything 3
~\citep{lin2025depthanything3}. TuringPAI-VL scores 19.99, compared with
2.03 for the strongest baseline, Qwen2-VL-2B Base.

The same checkpoint directly predicts 3D boxes on all four localization
benchmarks, with mAP ranging from 20.00 to 39.55 at a 3D IoU threshold of 0.15.
We use the Omni3D evaluator~\citep{brazil2023omni3d} and the Qwen3-VL
evaluation setup~\citep{bai2025qwen3vl} with confidence fixed at 1.0. None of
the reference models yields a valid score under this protocol.
TuringPAI-VL achieves the highest BLINK score among the three evaluated
models (49.50). On EmbSpatial, it scores 48.27, above InternVL3-1B Base
(39.86) and slightly below Qwen2-VL-2B Base (48.74).

\paragraph{Multi-image and temporal understanding.} We evaluate understanding
across sequences of frames on three video benchmarks
(Table~\ref{tab:vlm-1b}). TuringPAI-VL outperforms InternVL3-1B Base on all
three: MVBench (61.95 vs. 55.13), MLVU (51.50 vs. 47.21), and LongVideoBench
(45.60 vs. 40.24).

\begingroup

\begin{table}[!htbp]
  \caption{\textbf{Spatial perception and reasoning results for TuringPAI-VL and two
  public reference VLMs.} Bold and underlined mark the best and second-best
  valid scores among the displayed models.}
  \label{tab:spatial-perception}
  \centering
  \footnotesize
  \setlength{\tabcolsep}{3pt}
  \renewcommand{\arraystretch}{1.15}
  
  \begin{tabular}{llcc>{\columncolor{oursblue}}c}
    \toprule
    Category & Benchmark / Metric
      & \shortstack{Qwen2-VL-2B\\Base~\citep{wang2024qwen2vl}}
      & \shortstack{InternVL3-1B\\Base~\citep{zhu2025internvl3}}
      & TuringPAI-VL \\
    \midrule
    \multirow{3}{*}{\shortstack[l]{Geometric\\Perception}}
      & NYU Depth V2~\citep{silberman2012indoor} (\textit{AbsRel} $\downarrow$) & 0.70 & \underline{0.38} & \textbf{0.146} \\
      & ETH3D~\citep{schops2017eth3d} (EPE in px $\downarrow$) & 235.57 & \underline{199.05} & \textbf{193.49} \\
      & 7-Scenes~\citep{shotton2013scene} (AUC@30$^\circ$ $\uparrow$) & \underline{2.03} & 0.35 & \textbf{19.99} \\
    \cmidrule(lr){1-5}
    \multirow{4}{*}{\shortstack[l]{3D Object\\Localization}}
      & ARKitScenes~\citep{baruch2021arkitscenes} ($n=1{,}090$) & N/V & N/V & \textbf{33.76} \\
      & ScanNet~\citep{dai2017scannet} ($n=910$) & N/V & N/V & \textbf{39.55} \\
      & SUN RGB-D~\citep{song2015sunrgbd} ($n=1{,}859$) & N/V & N/V & \textbf{30.12} \\
      & nuScenes~\citep{caesar2020nuscenes} ($n=2{,}000$) & N/V & N/V & \textbf{20.00} \\
    \cmidrule(lr){1-5}
    \multirow{5}{*}{\shortstack[l]{Spatial\\Understanding}}
      & BLINK~\citep{fu2024blink} (score $\uparrow$) & \underline{42.30} & 42.14 & \textbf{49.50} \\
      & EmbSpatial~\citep{du2024embspatial} (score $\uparrow$) & \textbf{48.74} & 39.86 & \underline{48.27} \\
      & ERQA~\citep{gemini2025robotics} (score $\uparrow$) & \underline{32.5} & \textbf{33} & 30.0 \\
      & RoboSpatial~\citep{song2025robospatial} (score $\uparrow$) & \underline{38.0} & \textbf{40.0} & 30.0 \\
      & VSI~\citep{yang2025vsibench} (score $\uparrow$) & 20.60 & \textbf{31.37} & \underline{26.23} \\
    \bottomrule
  \end{tabular}%
  \par\vspace{2pt}
  \begin{minipage}{\linewidth}
    \footnotesize
    \textit{Note.} Arrows show metric direction. N/V denotes an attempted evaluation
    without a valid score; -- denotes unavailable scores. 3D localization is
    mAP at IoU 0.15; case counts are in parentheses.
  \end{minipage}
\end{table}
\endgroup

\subsection{TuringPAI-Drive: Autonomous Driving VLA Training and Evaluation}
\label{sec:exp-driving}

TuringPAI-Drive is initialized from the TuringPAI-VL checkpoint described in Section~\ref{sec:exp-vlm}, with a trajectory-planning head added for end-to-end trajectory prediction. This section presents a comprehensive evaluation of TuringPAI-Drive from four aspects. We first introduce the driving inputs and training data, covering the multi-camera temporal observations and large-scale in-house driving dataset. We then evaluate driving performance through open-loop trajectory prediction and XPeng's in-house CCES suite, with particular emphasis on comparing HybridViT and HybridViT-MLQ to assess the impact of MLQ-based visual token compression. Next, we present the on-device deployment and runtime optimizations enabled by XPeng Native Runtime (XNRT), focusing on inference efficiency and numerical accuracy on a single NVIDIA Orin-X SoC. Finally, we conduct an attribution analysis with qualitative visual comparisons between HybridViT and HybridViT-MLQ to further examine the effects of visual representation compression.

\subsubsection{Driving Inputs and Training Data}
\label{sec:exp-driving-setup}

TuringPAI-Drive takes four consecutive frames, including the current frame from a seven-camera surround-view system covering 360 degrees, sampled at 12\,Hz, as input and predicts a 6\,s ego trajectory represented by 72 waypoints.
The driving model uses the HybridViT-MLQ encoder described in Section~\ref{sec:mlq} to form a compact visual representation of the seven-camera input sequence. The model is initially trained on XPeng's in-house driving dataset,
comprising approximately 298,000 hours of recordings segmented into
35M clips. The data cover diverse driving environments, including
highways, urban roads, and parking scenarios. They encompass a broad
range of driving behaviors, including lateral maneuvers such as
automatic lane changes (ALC), navigation lane changes (NLC),
U-turns, and sharp curves; longitudinal maneuvers such as stopping
at red lights and starting from a standstill; and challenging
scenarios such as narrow-road negotiation, yielding to vulnerable
road users, and driver takeovers.

\subsubsection{Driving Evaluation and Results}
\label{sec:exp-driving-metrics}

The driving results in Tables~\ref{tab:vla-results} and~\ref{tab:vla-cces} are evaluated in PyTorch. Trajectory accuracy is measured by ADE and
FDE in metres, each split into lateral and longitudinal components and evaluated
over the full 6\,s horizon and the first 2\,s; mean per-point heading error
(yaw error, in radians) is also reported. Frames whose lateral endpoint error
at 6\,s exceeds 0.5\,m or whose longitudinal endpoint error exceeds 4.0\,m
count toward the miss rate.
Driving quality is evaluated on an in-house simulation set. Each CCES
axis---Compliance, Comfort, Efficiency, and Safety---aggregates
scenario-specific composite costs; Navigation is reported separately as a
route-following cost. Lower costs indicate better performance. The per-frame
collision-failure rate is the proportion of evaluation
frames flagged as failures by the collision detector.

\begin{table}[htbp]
  \caption{\textbf{Driving-VLA evaluation protocol.} Open-loop metrics are evaluated
  frame by frame on the 436k-frame validation split; driving-quality metrics
  are evaluated on an in-house simulation set. Lower values indicate better
  performance for all metrics.}
  \label{tab:vla-metrics-v2}
  \centering
  \small
  \setlength{\tabcolsep}{4pt}
  \renewcommand{\arraystretch}{1.15}
  \begin{tabularx}{\linewidth}{@{}p{0.18\linewidth}>{\raggedright\arraybackslash}p{0.19\linewidth}>{\raggedright\arraybackslash}X@{}}
    \toprule
    \textbf{Evaluation axis} & \textbf{Metric / suite} & \textbf{Definition} \\
    \midrule
    \multirow{3}{*}{\textbf{Open loop}}
      & ADE / FDE & Lateral and longitudinal displacement errors (m), over the 6\,s and 2\,s horizons \\
      & Yaw error & Mean per-point heading error (rad) \\
      & Miss rate & Frames whose lateral endpoint error at 6\,s exceeds 0.5\,m or whose longitudinal endpoint error exceeds 4.0\,m \\
    \midrule
    \multirow{5}{*}{\textbf{Driving quality}}
      & Compliance & Traffic-rule adherence, e.g., off-center driving and oncoming-lane entry \\
      & Comfort & e.g., acceleration jerk, swaying, and abnormal deceleration \\
      & Efficiency & e.g., speed-limit shortfall and slow acceleration \\
      & Safety & e.g., insufficient braking, roadside collision, and rear-collision risk from over-braking \\
    \cmidrule(lr){2-3}
      & Navigation & Route following, e.g., navigation-required lane changes, main/side-road splits, and intersections \\
    \bottomrule
  \end{tabularx}
\end{table}
\begin{table}[htbp]
  \caption{\textbf{Open-loop trajectory prediction performance on the held-out validation set.} Results are reported at $2$\,s and $6$\,s prediction horizons. ADE and FDE are measured in meters, yaw error in radians, and miss rate in percent. $\downarrow$ indicates that lower values are better, and bold denotes the best result in each column.}
  \label{tab:vla-results}
  \centering
  \small
  \setlength{\tabcolsep}{4pt}
  \renewcommand{\arraystretch}{1.22}
  \begin{tabular}{lcccccccccc}
    \toprule
    & \multicolumn{6}{c}{$6$\,s horizon} & \multicolumn{4}{c}{$2$\,s horizon} \\
    \cmidrule(lr){2-7}\cmidrule(lr){8-11}
    & \multicolumn{2}{c}{ADE (m) $\downarrow$} & \multicolumn{2}{c}{FDE (m) $\downarrow$} & \multicolumn{2}{c}{} & \multicolumn{2}{c}{ADE (m) $\downarrow$} & \multicolumn{2}{c}{FDE (m) $\downarrow$} \\
    \cmidrule(lr){2-3}\cmidrule(lr){4-5}\cmidrule(lr){8-9}\cmidrule(lr){10-11}
    Model & lat. & long. & lat. & long. & Yaw $\downarrow$ & Miss (\%) $\downarrow$ & lat. & long. & lat. & long. \\
    \midrule
    HybridViT & .1591 & \textbf{.9858} & .4076 & \textbf{2.7249} & .0129 & 37.58 & .0245 & \textbf{.1457} & .0686 & \textbf{.3573} \\
    \rowcolor{oursblue}\textbf{HybridViT-MLQ} & \textbf{.1572} & .9910 & \textbf{.4031} & 2.7345 & \textbf{.0128} & \textbf{37.48} & \textbf{.0243} & .1469 & \textbf{.0678} & .3605 \\
    \bottomrule
  \end{tabular}
\end{table}

\label{sec:exp-driving-openloop}

\paragraph{Open-loop trajectory prediction.}
We evaluate trajectory prediction accuracy using Average
Displacement Error (ADE) and Final Displacement Error (FDE),
measured separately in the lateral and longitudinal directions
over the full 6\,s horizon and the first 2\,s. We also report
mean per-point heading error (yaw error, in radians) and miss
rate, defined as the proportion of frames whose 6\,s endpoint
error exceeds 0.5\,m laterally or 4.0\,m longitudinally.

We evaluate the effect of MLQ-based token compression by comparing HybridViT-MLQ under the on-device token budget with the full-token HybridViT. At the $6$\,s horizon, HybridViT-MLQ achieves 1.2\% lower lateral ADE and 1.1\% lower lateral FDE, while incurring only marginal increases of 0.5\% and 0.4\% in longitudinal ADE and FDE, respectively. Across both prediction horizons, all ADE and FDE metrics remain within 1.2\% of those of the full-token HybridViT. Moreover, yaw error and miss rate are slightly reduced. These results demonstrate that MLQ effectively compresses the visual-token representation while largely preserving, and in some metrics slightly improving, trajectory prediction accuracy (Table~\ref{tab:vla-results}).

\paragraph{Driving quality.}
We further evaluate driving quality using XPeng's in-house
simulation suite. The evaluation covers four CCES dimensions:
Compliance, Comfort, Efficiency, and Safety, with Navigation
assessed separately through route-following costs. For each
metric, we report the ratio of its original, unnormalized cost
to that of the full-token HybridViT reference (1.0). Category-level
scores are computed as the geometric mean of the corresponding
per-metric ratios, with lower values indicating better performance.
We additionally report the collision-failure rate as the percentage
of evaluation frames flagged by the collision detector.

\begin{table}[t]
  \caption{\textbf{Per-metric ratios of unnormalized driving-quality costs, with
  HybridViT as the reference (1.00); lower ratios indicate lower costs for
  HybridViT-MLQ.} Bold indicates ratios below 1.0. Each Avg. is the geometric mean over
  all subclasses in its category, not only those shown. Navigation reports the
  aggregate route-following cost.}
  \label{tab:vla-cces}
  \centering
  \small
  \setlength{\tabcolsep}{2.5pt}
  \renewcommand{\arraystretch}{1.12}
  \begin{tabularx}{\linewidth}{@{\hspace{\tabcolsep}}*{10}{>{\centering\arraybackslash}X}@{\hspace{\tabcolsep}}}
    \toprule
    \multicolumn{5}{c}{Compliance $\downarrow$} & \multicolumn{4}{c}{Comfort $\downarrow$} &
    Navi.~$\downarrow$ \\
    \cmidrule(lr){1-5}\cmidrule(lr){6-9}\cmidrule(lr){10-10}
    Off-Ctr. & Onc.~Ln. & Tra.~Lig. & Line~X & Avg. & Jerk & Sway & Decel. & Avg. &
    Avg. \\
    \midrule
    \rowcolor{oursblue} 1.009 & 1.010 & \textbf{0.997} & 1.004 & 1.002 & 1.061 &
    \textbf{0.977} & \textbf{0.976} & 1.004 &
    \textbf{0.991} \\
    \midrule
    \addlinespace[4pt]
    \multicolumn{5}{c}{Efficiency $\downarrow$} & \multicolumn{5}{c}{Safety $\downarrow$} \\
    \cmidrule(lr){1-5}\cmidrule(lr){6-10}
    Sl.~Acc. & Spd.~Sh. & Late~LC & Long~F. & Avg. & Ins.~Brk. & Roadside & Risky~LC & Rear~Risk & Avg. \\
    \midrule
    \rowcolor{oursblue} \textbf{0.992} & \textbf{0.973} & \textbf{0.996} & 1.003 &
    \textbf{0.992} & \textbf{0.998} & 1.015 & \textbf{0.997} &
    \textbf{0.962} & \textbf{0.995} \\
    \bottomrule
  \end{tabularx}

  \vspace{2pt}
  \begin{minipage}{\linewidth}
    \footnotesize
    Abbreviations: Navi., Navigation (aggregate route-following cost); Off-Ctr., off-center driving; Onc. Ln., oncoming-lane entry;
    Tra. Lig., traffic-light issues; Line X, line crossing; Decel., abnormal
    deceleration; Sl. Acc., slow acceleration; Spd. Sh., speed-limit shortfall;
    Late LC, delayed lane change; Long F., long following distance; Ins. Brk.,
    insufficient braking; Roadside, roadside collision; Risky LC, risky lane change;
    Rear Risk, rear-collision risk. Jerk, acceleration jerk; Sway, swaying.
  \end{minipage}
\end{table}

As shown in Table~\ref{tab:vla-cces}, HybridViT-MLQ
maintains comparable driving quality to HybridViT, with all
five category-level ratios within 0.9\% of the reference:
Safety (0.995), Comfort (1.004), Efficiency (0.992),
Compliance (1.002), and Navigation (0.991). At the
per-metric level, HybridViT-MLQ reduces costs associated with
rear collision risk (0.962), abnormal deceleration (0.976),
speed-limit shortfall (0.973), and swaying (0.977), while
showing small regressions in acceleration jerk (1.061),
roadside collision (1.015), oncoming-lane entry (1.010),
and off-center driving (1.009). The collision-failure rate
increases slightly from 0.258\% to 0.263\%.
Overall, these results indicate that the compact visual
representation preserves aggregate driving quality, although
some individual metrics exhibit trade-offs (Table~\ref{tab:vla-cces}).
A full per-metric breakdown is provided in Appendix~\ref{app:cces-breakdown}
(Table~\ref{app:cces-full}).

\subsubsection{On-Device Deployment and Runtime Optimization}
TuringPAI-Drive is XPeng's VLA 2.0 Lite model and uses a 0.65B VLM backbone. XPeng's NVIDIA Orin-X-based in-vehicle computing platform contains two Orin-X SoCs; TuringPAI-Drive deployment and all latency measurements reported here use a single Orin-X SoC. The GPU and DLA stages of the model execute within that same SoC. Deploying TuringPAI-Drive on automotive-grade hardware presents significant challenges in balancing inference latency, numerical accuracy, and computational efficiency. 
This optimization reduces the GPU-side VLM prefill latency to 69.773 ms, bringing the VLM backbone below the 83.3 ms reference period corresponding to 12 Hz. At the system level, TuringPAI-Drive generates trajectories at 12\,Hz in on-vehicle operation.

To address these challenges, we deploy TuringPAI-Drive using the XPeng Native Runtime (XNRT), an in-house inference runtime specifically designed for efficient on-device AI deployment. Unlike conventional deployment frameworks that primarily rely on general-purpose operator optimization, XNRT provides a unified optimization framework across model architecture, numerical precision, and target hardware. It is designed to be quantization-aware-training (QAT)-friendly and supports model-specific, precision-aware, and hardware-aware optimizations. Through these capabilities, XNRT enables aggressive computational acceleration while preserving numerical fidelity, making it possible to deploy TuringPAI-Drive on a single Orin-X SoC within the required real-time latency budget.

\paragraph{Evaluation Setup.}
We evaluate the ViT and LLM components of TuringPAI-Drive on a single NVIDIA Orin-X SoC. We compare two deployment frameworks, NVIDIA TensorRT~8.6 and XPeng XNRT, under four configurations: TensorRT FP32, TensorRT FP32+INT8 mixed precision with QAT, TensorRT FP16, and XNRT BF16+INT8 with QAT.
Reported VLM prefill inference latency is the sum of the GPU-side ViT and LLM inference times, excluding preprocessing, data transfers, DLA execution, and the action head.
Numerical fidelity is evaluated on real-world data against reference outputs produced by the original model in PyTorch FP32.

The two INT8 configurations use QAT and SmoothQuant to mitigate quantization-induced numerical degradation, while differing in quantization granularity and execution precision.
TensorRT FP32 and FP16 serve as additional precision baselines.

For the TensorRT FP32+INT8 configuration, per-tensor INT8 quantization is applied to Linear and Convolution operators, while the remaining operators primarily execute in FP32.
The FP16 configuration exhibits substantial numerical degradation, as reflected by its low cosine similarity and extremely large maximum absolute error. Therefore, FP32+INT8 mixed precision is adopted as the TensorRT mixed-precision baseline.

In contrast, XNRT adopts a more flexible mixed-precision execution strategy. It applies dynamic per-token INT8 quantization to Linear and Convolution operators, while executing most remaining operators in BF16. FP32 precision is retained for numerically sensitive operations, including LayerNorm, RMSNorm, and operators involving transcendental functions such as Tanh. Beyond quantization, XNRT further incorporates operator fusion and dimension-aware custom optimizations to reduce memory-access overhead and kernel scheduling costs. This combination of fine-grained quantization, mixed-precision execution, and hardware-specific operator optimization enables XNRT to achieve substantially higher inference efficiency while maintaining high numerical fidelity. In particular, XNRT preserves mean and minimum cosine similarities of 0.998369 and 0.998106, respectively, relative to the PyTorch FP32 reference, substantially outperforming the TensorRT FP32+INT8 configuration.

We evaluate deployment performance using VLM prefill inference latency and numerical fidelity metrics, including cosine mean, cosine minimum, mean absolute error (MAE), root mean squared
error (RMSE), worst-case maximum absolute error, and mean relative $L_2$ error.

\begin{figure}[t] %
  \centering
  \includegraphics[width=\textwidth]{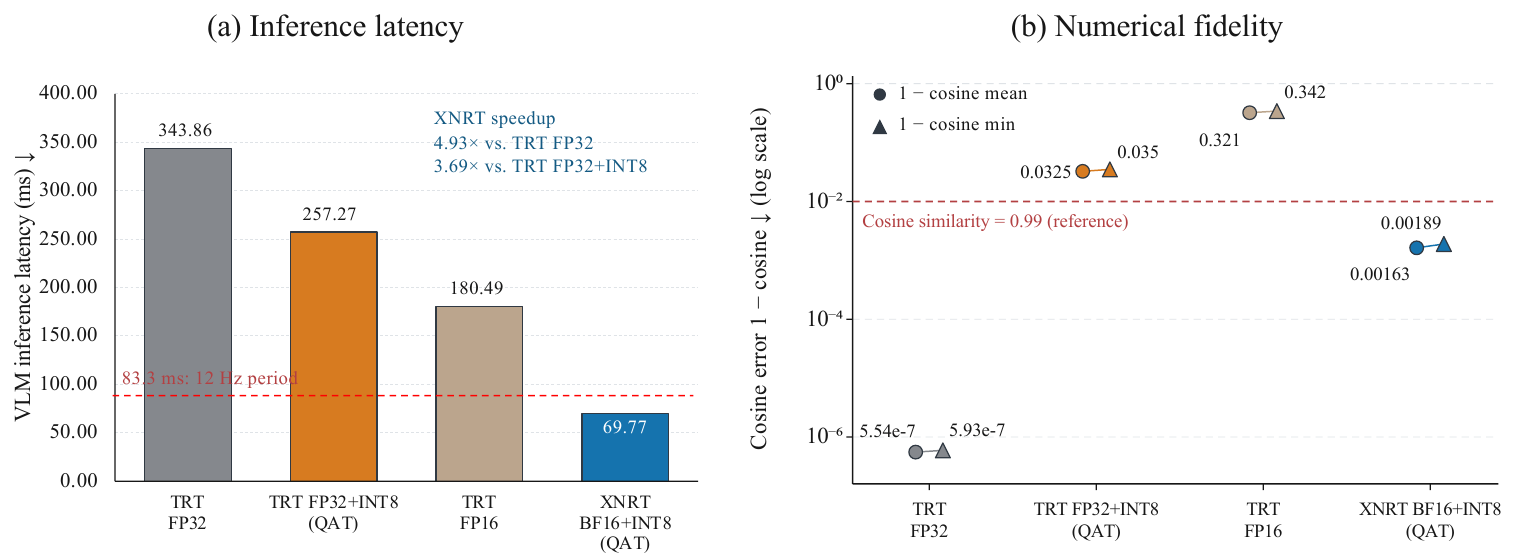}
  \caption{\textbf{On-device latency and numerical fidelity of the
TuringPAI-Drive VLM backbone on a single NVIDIA Orin-X SoC.}
(a) GPU-side VLM prefill inference latency across four deployment
configurations. The dashed line marks the 83.3\,ms period
corresponding to 12\,Hz.
(b) Cosine errors derived from mean and minimum cosine similarities,
shown on a logarithmic scale. The dashed line corresponds to
cosine similarity 0.99. Lower values are better in both panels.
TRT denotes TensorRT~8.6.}

  \label{fig:deployment}
\end{figure}

\paragraph{Latency and Accuracy Evaluation.}
Table~\ref{tab:deployment-comparison} and  Fig.~\ref{fig:deployment} summarize the comparison across the four deployment configurations.

\begin{table}[htbp]
  \centering
  \caption{\textbf{VLM inference latency and numerical fidelity on a single NVIDIA Orin-X SoC.} Latency sums GPU-side ViT and LLM times, excluding preprocessing, transfers, DLA execution, and the action head. Outputs are compared with the original model's PyTorch FP32 reference. QAT: quantization-aware training; --: unreported. Best values are bold.}
  \label{tab:deployment-comparison}
  \begingroup
  \footnotesize
  \setlength{\tabcolsep}{4pt}
  \renewcommand{\arraystretch}{1.0}
  \resizebox{\linewidth}{!}{%
  \begin{tabular}{@{}lrrrrrrr@{}}
    \toprule
    & Latency & Cosine & Cosine & MAE & RMSE & Max. abs. & Relative $L_2$ \\
    Configuration & (ms)$\downarrow$ & mean$\uparrow$ & min$\uparrow$ & mean$\downarrow$ & mean$\downarrow$ & worst$\downarrow$ & mean$\downarrow$ \\
    \midrule
    TensorRT FP32 & 343.863 & \textbf{0.999999} & \textbf{0.999999} & \textbf{0.000631} & \textbf{0.001150} & \textbf{0.102601} & \textbf{0.001068} \\
    TensorRT FP32+INT8 (QAT) & 257.270 & 0.967537 & 0.964966 & 0.155759 & 0.276348 & 24.122121 & 0.253196 \\
    TensorRT FP16 & 180.489 & 0.678869 & 0.657518 & 0.568197 & 0.852783 & 125.000000 & -- \\
    \rowcolor{oursblue} XNRT BF16+INT8 (QAT) & \textbf{69.773} & 0.998369 & 0.998106 & 0.035446 & 0.062417 & 6.312500 & 0.057103 \\
    \bottomrule
  \end{tabular}%
  }
  \endgroup
\end{table}

TensorRT FP32 achieves a VLM prefill latency of 343.863 ms and a cosine mean of 0.999999, providing a near-lossless reference for numerical accuracy.

\begin{samepage}
\noindent Introducing FP32+INT8 mixed precision reduces the latency to 257.270 ms. However, its cosine mean and cosine minimum decrease substantially to 0.967537 and 0.964966, respectively, both falling below the 0.99 reference shown in Fig.~\ref{fig:deployment}(b), indicating reduced numerical fidelity.
\par
\end{samepage}

TensorRT FP16 further reduces latency to 180.489\,ms, but exhibits substantially larger numerical deviations. Its cosine mean and minimum are 0.678869 and 0.657518, respectively, and its worst-case maximum absolute error reaches 125.0. These results show that lower execution precision alone does not provide a satisfactory balance between latency and numerical fidelity for this model.

In comparison, XNRT BF16+INT8 achieves a VLM prefill latency of only 69.773 ms, while maintaining a cosine mean of 0.998369 and a cosine minimum of 0.998106. Both metrics remain above 0.99, and the cosine mean is only 0.001630 below that of the TensorRT FP32 baseline.
Importantly, XNRT's VLM prefill latency is approximately 13.5\,ms below the 83.3\,ms period corresponding to 12\,Hz. This places the measured VLM stage within the target execution
budget while maintaining high numerical fidelity.

In terms of inference efficiency, XNRT achieves a $4.93\times$ speedup over TensorRT FP32, corresponding to a 79.71\% reduction in VLM prefill latency. Compared with TensorRT FP32+INT8 mixed precision, XNRT delivers a further $3.69\times$ speedup, reducing latency by 72.88\%. These improvements demonstrate the effectiveness of XNRT's combined optimization strategy, particularly its dynamic per-token quantization, BF16 execution for non-GEMM operators, operator fusion, and hardware-aware kernel optimization.

The accuracy comparison further highlights the advantage of XNRT. Relative to TensorRT FP32+INT8 mixed precision, XNRT improves cosine mean by 0.030832, reducing the cosine error ($1-\text{cosine mean}$) to approximately $1/19.9$ of that of the TensorRT configuration. Meanwhile, its MAE, RMSE, maximum absolute error, and relative $L_2$ error are approximately $4.39\times$, $4.43\times$, $3.82\times$, and $4.43\times$ lower, respectively. These results indicate that the latency reduction achieved by XNRT does not come at the expense of substantial numerical accuracy loss. Instead, its fine-grained quantization and heterogeneous precision execution provide a more favorable balance between computational efficiency and numerical fidelity.

\paragraph{Deployment Significance.}
Taken together, XNRT establishes a substantially improved latency--accuracy Pareto frontier for on-device VLA inference on a single Orin-X SoC. While TensorRT FP32 provides the highest numerical accuracy, its 343.863 ms latency is approximately 4.93 times that of XNRT and is far beyond the 12 FPS requirement. TensorRT FP32+INT8 mixed precision reduces latency but suffers from considerable accuracy degradation. In contrast, XNRT achieves both a sub-70 ms VLM prefill latency and cosine accuracy metrics above 0.99, outperforming the TensorRT mixed-precision configuration in both latency and numerical accuracy.

These results establish XNRT as a key enabling component for deploying large-scale VLA models on resource-constrained automotive platforms. By jointly optimizing quantization granularity, numerical precision, operator execution, and hardware-specific scheduling, XNRT makes it feasible to run TuringPAI-Drive on a single NVIDIA Orin-X SoC while meeting the real-time requirements of XPeng's Max intelligent vehicles. These vehicles use a two-SoC Orin-X computing platform, whereas the model deployment and reported evaluation use one Orin-X SoC. Future work will further investigate the generalization of these optimizations across different vehicle platforms, input distributions, and VLA model scales.

\subsubsection{Attribution Analysis}
\label{sec:exp-driving-attribution}
\begin{figure}[htbp]
  \centering
  \newcommand{\figSevenCamHeader}[2]{%
    \begin{minipage}[b]{#1}\centering\scriptsize #2\end{minipage}%
  }
  \newcommand{\figSevenCamImage}[2]{%
    \begin{minipage}[c]{#1}\centering#2\end{minipage}%
  }
  \newcommand{\figSevenCamRowLabel}[1]{%
    \begin{minipage}[c][1.17cm][c]{0.5cm}\centering\rotatebox{90}{\fontsize{4.75}{5.7}\selectfont #1}\end{minipage}%
  }
  \newcommand{\figSevenCamRow}[8]{%
    \par\noindent\makebox[\linewidth][c]{%
      \figSevenCamRowLabel{#1}%
      \figSevenCamImage{1.755cm}{#2}%
      \figSevenCamImage{1.560cm}{#3}%
      \figSevenCamImage{2.048cm}{#4}%
      \figSevenCamImage{2.340cm}{#5}%
      \figSevenCamImage{2.340cm}{#6}%
      \figSevenCamImage{1.560cm}{#7}%
      \figSevenCamImage{1.755cm}{#8}%
    }\par\vspace{0.8mm}%
  }
  \makebox[\linewidth][c]{%
    \makebox[0.5cm]{}%
    \figSevenCamHeader{1.755cm}{Left rear}%
    \figSevenCamHeader{1.560cm}{Left front}%
    \figSevenCamHeader{2.048cm}{Front narrow}%
    \figSevenCamHeader{2.340cm}{Front fisheye}%
    \figSevenCamHeader{2.340cm}{Rear}%
    \figSevenCamHeader{1.560cm}{Right front}%
    \figSevenCamHeader{1.755cm}{Right rear}%
  }\par\vspace{0.8mm}
  \figSevenCamRow
    {Camera}%
    {\includegraphics[height=1.17cm]{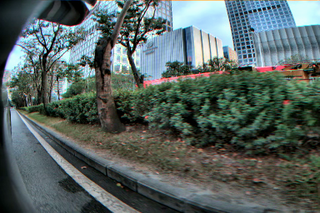}}%
    {\includegraphics[height=1.17cm]{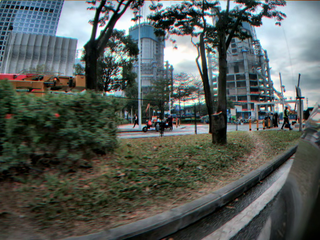}}%
    {\includegraphics[height=1.17cm]{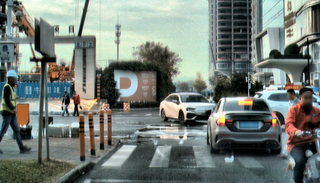}}%
    {\includegraphics[height=1.17cm]{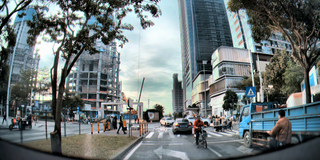}}%
    {\includegraphics[height=1.17cm]{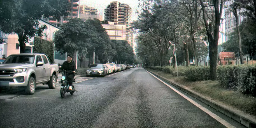}}%
    {\includegraphics[height=1.17cm]{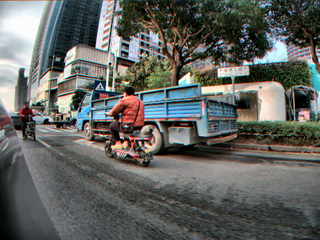}}%
    {\includegraphics[height=1.17cm]{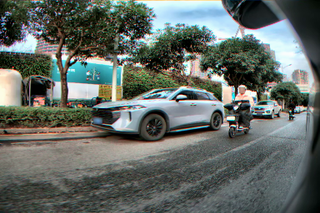}}

  \figSevenCamRow
    {HybridViT}%
    {\includegraphics[height=1.17cm]{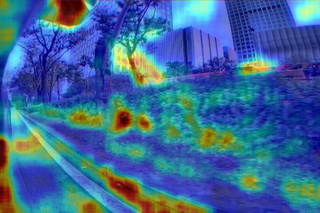}}%
    {\includegraphics[height=1.17cm]{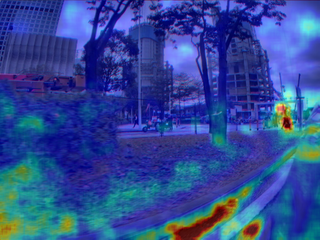}}%
    {\includegraphics[height=1.17cm]{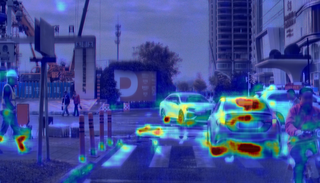}}%
    {\includegraphics[height=1.17cm]{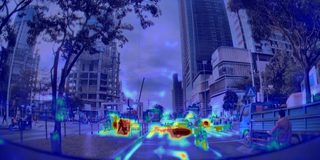}}%
    {\includegraphics[height=1.17cm]{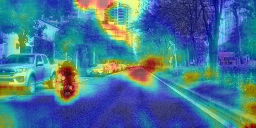}}%
    {\includegraphics[height=1.17cm]{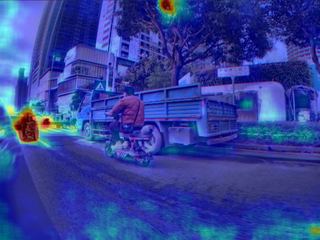}}%
    {\includegraphics[height=1.17cm]{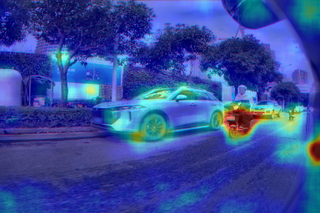}}

  \figSevenCamRow
    {HybridViT-MLQ}%
    {\includegraphics[height=1.17cm]{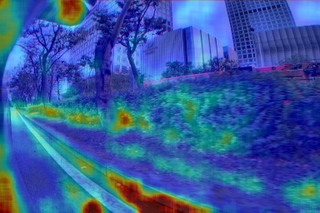}}%
    {\includegraphics[height=1.17cm]{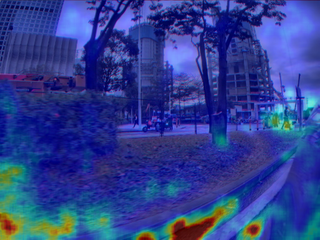}}%
    {\includegraphics[height=1.17cm]{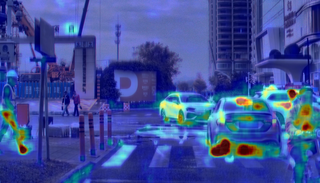}}%
    {\includegraphics[height=1.17cm]{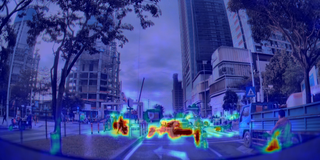}}%
    {\includegraphics[height=1.17cm]{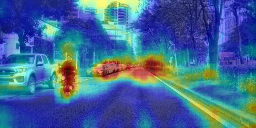}}%
    {\includegraphics[height=1.17cm]{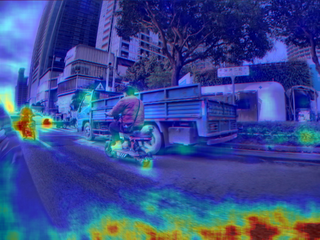}}%
    {\includegraphics[height=1.17cm]{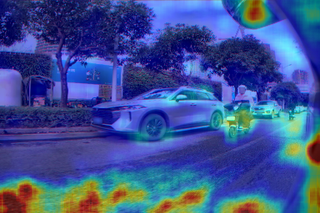}}

  {\normalsize (a) Daytime left turn}\par\vspace{0.5mm}
  \figSevenCamRow
    {Camera}%
    {\includegraphics[height=1.17cm]{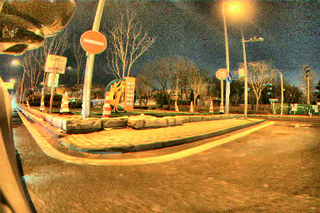}}%
    {\includegraphics[height=1.17cm]{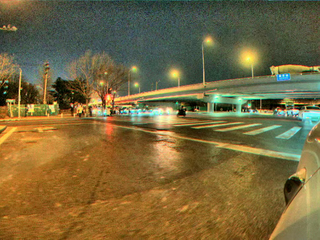}}%
    {\includegraphics[height=1.17cm]{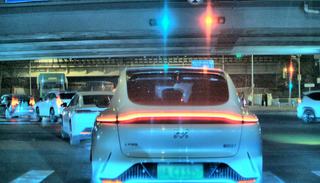}}%
    {\includegraphics[height=1.17cm]{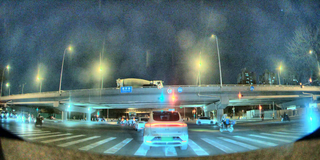}}%
    {\includegraphics[height=1.17cm]{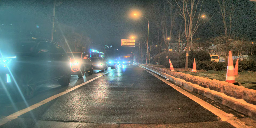}}%
    {\includegraphics[height=1.17cm]{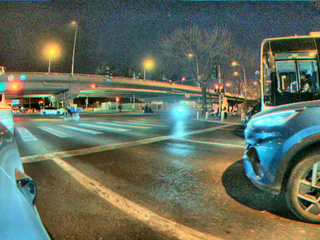}}%
    {\includegraphics[height=1.17cm]{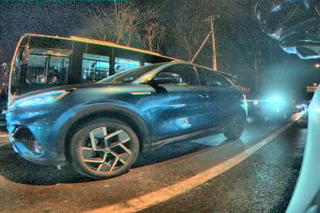}}

  \figSevenCamRow
    {HybridViT}%
    {\includegraphics[height=1.17cm]{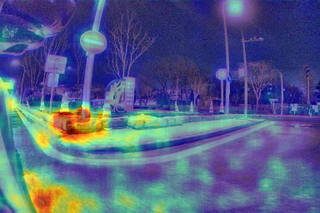}}%
    {\includegraphics[height=1.17cm]{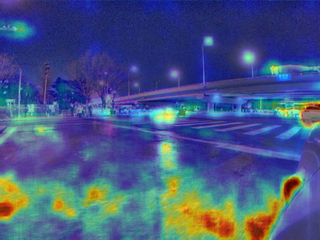}}%
    {\includegraphics[height=1.17cm]{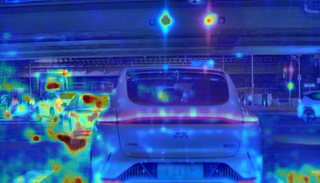}}%
    {\includegraphics[height=1.17cm]{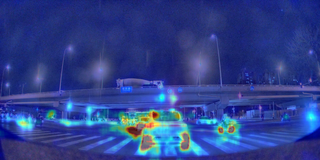}}%
    {\includegraphics[height=1.17cm]{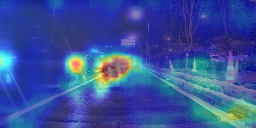}}%
    {\includegraphics[height=1.17cm]{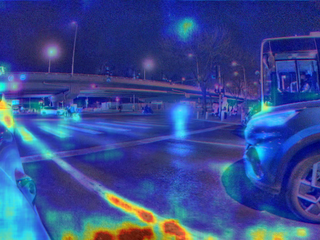}}%
    {\includegraphics[height=1.17cm]{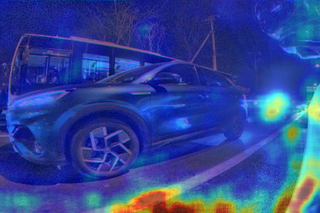}}

  \figSevenCamRow
    {HybridViT-MLQ}%
    {\includegraphics[height=1.17cm]{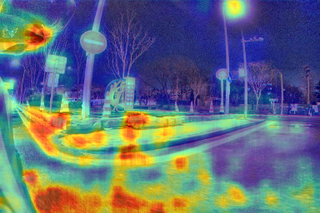}}%
    {\includegraphics[height=1.17cm]{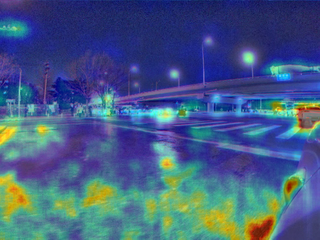}}%
    {\includegraphics[height=1.17cm]{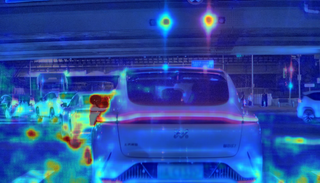}}%
    {\includegraphics[height=1.17cm]{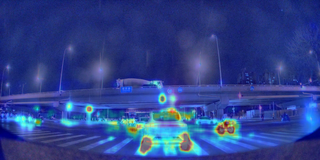}}%
    {\includegraphics[height=1.17cm]{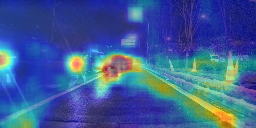}}%
    {\includegraphics[height=1.17cm]{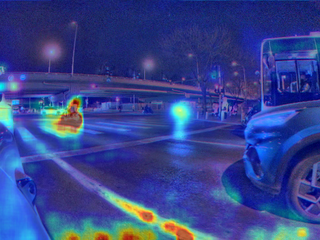}}%
    {\includegraphics[height=1.17cm]{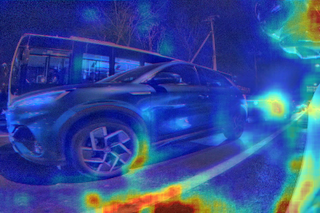}}

  {\normalsize (b) Nighttime left turn}\par\vspace{0.5mm}
  \caption{\textbf{Integrated Gradients attribution maps for the planned trajectory
  across seven camera views in two driving scenes: (a) a daytime left turn and
  (b) a nighttime left turn.} For each scene, rows from top to bottom show the
  original camera images, HybridViT attributions, and HybridViT-MLQ attributions;
  camera views are labeled above. The visualizations correspond to the final
  frame of the four-frame input sequence.}
  \label{fig:vla-saliency}
\end{figure}

To complement the quantitative driving evaluation above, we conduct a qualitative
analysis of the visual evidence associated with the planned trajectory in HybridViT and HybridViT-MLQ, asking whether HybridViT-MLQ's efficient, multi-scale aggregation of visual evidence retains the key visual cues associated with the planned trajectory. Attribution maps (or saliency maps) are a standard post-hoc tool for explaining vision models~\citep{deng2025attribution, itti2001computational, simonyan2013deep, adebayo2018sanity}, but have been less explored in driving-VLA evaluation~\citep{jiang2025vla4ad}. We apply Integrated Gradients, which attributes a model output to its input features by integrating gradients along a straight-line path from a baseline input to the observed input~\citep{sundararajan2017integrated}. Specifically, we compute pixel-level attributions with respect to the planned trajectory and visualize them for all seven camera views at the last frame of the input
sequence. Figure~\ref{fig:vla-saliency} presents two representative scenes; for each
scene, the rows show the original images, followed by the HybridViT and
HybridViT-MLQ attribution maps.

Across the cases we examined, HybridViT and HybridViT-MLQ assign attribution to broadly 
similar objects at similar image locations and HybridViT-MLQ  tends to produce
more spatially concentrated attribution around key objects, with weaker
responses over background regions such as the sky, buildings, and
vegetation, although this pattern varies across camera views. In Figure
~\ref{fig:vla-saliency}(a), both models highlight the car directly ahead,
the cyclist on the right, and the pedestrian at the front left near the
crossing. In Figure~\ref{fig:vla-saliency}(b), both highlight the traffic
signal, the rear wheel or taillights of the lead SUV, while showing weaker responses in the darker
surroundings. In the front fisheye view, HybridViT-MLQ also shows attribution around a front-left VRU 
that is less apparent in the HybridViT maps; this may indicate an additional
visual cue associated with the planned trajectory. This is a qualitative difference 
in the attribution maps; further examples are provided in Appendix~\ref{app:saliency-cases}.

\clearpage
\subsection{TuringPAI-Embodied: Robotic Manipulation VLA}
\label{sec:exp-embodied}

TuringPAI-Embodied combines the pretrained TuringPAI-VL backbone from
Section~\ref{sec:exp-vlm} with a $\pi_0$-style flow-matching action
expert~\citep{black2024pi0}. Each observation includes two $256\times256$
camera views and an 80-dimensional canonical robot state. The action expert
uses ten denoising steps to generate a chunk of six consecutive actions in a
canonical action representation. We pretrain the policy on roughly 500 hours
of manipulation data from multiple robot embodiments and then fine-tune it
with supervision from LIBERO demonstrations. We use the resulting policy for
the LIBERO evaluation and the on-device timing measurements below.

TuringPAI-Embodied succeeded in 1,936 of 2,000 LIBERO~\citep{liu2023libero} trials
(96.8\%), with 500 trials per suite (Table~\ref{tab:emb-libero}).
Per-suite success rates range from 93.2\% on LIBERO-10 to 99.6\% on Object.
The overall rate is close to the 96.85\% OpenPI reports for its
LIBERO-fine-tuned $\pi_{0.5}$ policy~\citep{black2025pi05,openpiLiberoResults}.

\begin{table}[htbp]
  \caption{\textbf{LIBERO success rates (\%) by suite and overall.}
  TuringPAI-Embodied was evaluated on 500 trials per suite. The $\pi_{0.5}$ scores
  are the published results for the OpenPI \texttt{pi05\_libero}
  checkpoint~\citep{openpiLiberoResults}.}
  \label{tab:emb-libero}
  \centering
  \small
  \setlength{\tabcolsep}{6pt}
  \begin{tabular}{lccccc}
    \toprule
    Model & Spatial & Object & Goal & LIBERO-10 & Overall \\
    \midrule
    OpenPI $\pi_{0.5}$ (published) & 98.8 & 98.2 & 98.0 & 92.4 & 96.85 \\
    \rowcolor{oursblue}TuringPAI-Embodied (ours) & 98.4 & 99.6 & 96.0 & 93.2 & 96.8 \\
    \bottomrule
  \end{tabular}
\end{table}

\begin{table}[htbp]
  \caption{\textbf{Reference inference-stage timings (ms) on a single NVIDIA Orin-X SoC.}
  TuringPAI-Embodied can run CNN patch embedding on the DLA. To compare
  the subsequent stages, vision timing starts after patch embedding for
  both policies. These timings are provided for reference only.}
  \label{tab:emb-latency}
  \centering
  \small
  \setlength{\tabcolsep}{6pt}
  \begin{tabular}{lrrrr}
    \toprule
    Policy & Vision & LLM & Action expert & Sum \\
    \midrule
    OpenPI $\pi_{0.5}$(published) & 84.1 & 220.6 & 117.5 & 422.2 \\
    \rowcolor{oursblue} TuringPAI-Embodied (ours) & 27.3 & 60.8 & 80.2 & 168.3 \\
    \bottomrule
  \end{tabular}
\end{table}

TuringPAI-Embodied's CNN patch embedding can run on the DLA, allowing it
to overlap with GPU work from another inference cycle in a pipelined
deployment. To isolate the stages that follow this front end, we start
vision timing after patch embedding and use the same timing boundary for
the $\pi_{0.5}$ reference (Table~\ref{tab:emb-latency}). We profile both
policies on a single NVIDIA Orin-X SoC in MAXN mode using their respective
LIBERO input configurations. Both LLMs run in PyTorch eager mode, and both
action experts use CUDA Graph replay for ten denoising steps. The measured
vision, LLM, and action-expert times sum to 168.3\,ms for
TuringPAI-Embodied and 422.2\,ms for $\pi_{0.5}$.

Asynchronous action execution can overlap prediction of the next chunk with
execution of the current one. The predicted actions can then be based on
observations acquired before the intended execution time, creating
perception--execution misalignment~\citep{yang2026eaglevla}. Future work will
evaluate asynchronous action execution for TuringPAI-Embodied, drawing on
approaches such as VLASH~\citep{tang2025vlash} and RTC~\citep{black2026real}.

\section{Conclusion and Future Work}
\label{sec:conclusion}

In this work, we presented TuringPAI, a foundation for on-device
Physical AI that connects general vision-language understanding
with real-world autonomous driving and robotic intelligence.
At its core, TuringPAI-VL combines HybridViT-MLQ with
Physical-AI-oriented training to develop general multimodal,
spatial, geometric, and temporal understanding within a compact
0.65B-parameter model. By integrating convolutional local
aggregation with multi-scale latent queries, HybridViT-MLQ
substantially reduces the visual-token budget while preserving
the spatial and temporal evidence required by downstream
physical tasks. Building on this foundation, we developed two domain-specific
VLA systems, TuringPAI-Drive and TuringPAI-Embodied, demonstrating
the transfer of shared vision-language capabilities to autonomous
driving and robotic manipulation. TuringPAI-Drive preserves
trajectory prediction accuracy within 1.2\% of the full-token
HybridViT baseline and generates trajectories at 12\,Hz on a single NVIDIA
Orin-X SoC in on-vehicle operation. TuringPAI-Embodied achieves a 96.8\% average success rate
across the four LIBERO suites. On a single NVIDIA Orin-X SoC, the measured
inference stages after CNN patch embedding sum to 168.3\,ms for
TuringPAI-Embodied. Together, these results
demonstrate the feasibility of a shared Physical AI foundation
supporting distinct driving and robotic systems under practical
edge-computing constraints.

Looking ahead, we see several opportunities to advance TuringPAI
toward more general and autonomous Physical AI. First, we aim to
extend temporal modeling beyond short observation windows toward
long-horizon physical understanding, incorporating persistent
temporal memory, streaming representations, and action-conditioned
future prediction to model evolving environments and interaction
dynamics. Second, we plan to strengthen the embodied foundation
through large-scale cross-embodiment and cross-task learning,
incorporating richer proprioceptive and contact information to
improve generalization across robot platforms and interaction
scenarios. Finally, we aim to develop closed-loop Physical AI
systems that integrate world models, outcome evaluation, and
failure-aware recovery, enabling policies to predict, act, evaluate,
and adapt through sustained real-world interaction. Together with
continued advances in asynchronous inference and heterogeneous
edge computing, these directions will move TuringPAI toward
more capable, adaptive, and reliable Physical AI systems.


\section{Contributors}
\textbf{Advisors}: Hang Zhang, Honggou Yang, Xianming Liu

\textbf{Project Lead}: Qiman Wu

\textbf{Contributors}: 
Hanlin Chen,
Yuhua Wei,
Mingyuan Wang,
Shaman Tang,
Minghao Li,
Rui Xin,
Xingye Liu,
Zhichao Yang,
Yiqi Yu,
Hongyan Pei,
Xiaowen Tian,
YingYing Liu,
Shiwen Liu,
Jiahui Hu,
Yizhao Wang,
Yuheng Zhang,
Anhua Liu,
Yue He,
Shuang Su,
Da Zhu,
Hua Zhou,
Pengfei Diao,
Haojie Yang,
Weichao Huang

\bibliographystyle{unsrtnat}
\begingroup
\setlength{\bibsep}{1pt}
\bibliography{refs}

\begin{thebibliography}{75}
\providecommand{\natexlab}[1]{#1}
\providecommand{\url}[1]{\texttt{#1}}
\expandafter\ifx\csname urlstyle\endcsname\relax
  \providecommand{\doi}[1]{doi: #1}\else
  \providecommand{\doi}{doi: \begingroup \urlstyle{rm}\Url}\fi

\bibitem[Shukor et~al.(2025)Shukor, Aubakirova, Capuano, Kooijmans, Palma,
  Zouitine, Aractingi, Pascal, Russi, Marafioti, et~al.]{shukor2025smolvla}
Mustafa Shukor, Dana Aubakirova, Francesco Capuano, Pepijn Kooijmans, Steven
  Palma, Adil Zouitine, Michel Aractingi, Caroline Pascal, Martino Russi,
  Andres Marafioti, et~al.
\newblock {SmolVLA: A Vision-Language-Action Model for Affordable and Efficient
  Robotics}.
\newblock \emph{arXiv preprint arXiv:2506.01844}, 2025.

\bibitem[Black et~al.(2025)Black, Galliker, and Levine]{black2026real}
Kevin Black, Manuel Galliker, and Sergey Levine.
\newblock Real-time execution of action chunking flow policies.
\newblock \emph{Advances in Neural Information Processing Systems},
  38:\penalty0 33383--33407, 2025.
\newblock \doi{10.52202/085713-1122}.

\bibitem[Zhang et~al.(2026)Zhang, Wang, Zhu, Zhou, He, Hu, Tang, Chen, Wei,
  Liu, et~al.]{zhang2026turingllm}
Yuheng Zhang, Yizhao Wang, Da~Zhu, Hua Zhou, Yue He, Jiahui Hu, Shaman Tang,
  Hanlin Chen, Yuhua Wei, Anhua Liu, et~al.
\newblock {TuringLLM: Efficiently Scaling Foundation Models Toward Physical
  AI}.
\newblock \emph{arXiv preprint arXiv:2608.30567}, 2026.

\bibitem[Liu et~al.(2023{\natexlab{a}})Liu, Li, Wu, and Lee]{liu2023visual}
Haotian Liu, Chunyuan Li, Qingyang Wu, and Yong~Jae Lee.
\newblock Visual instruction tuning, 2023{\natexlab{a}}.
\newblock URL \url{https://arxiv.org/abs/2304.08485}.

\bibitem[Wang et~al.(2024{\natexlab{a}})Wang, Bai, Tan, Wang, Fan, Bai, Chen,
  Liu, Wang, Ge, Fan, Dang, Du, Ren, Men, Liu, Zhou, Zhou, and
  Lin]{wang2024qwen2vl}
Peng Wang, Shuai Bai, Sinan Tan, Shijie Wang, Zhihao Fan, Jinze Bai, Keqin
  Chen, Xuejing Liu, Jialin Wang, Wenbin Ge, Yang Fan, Kai Dang, Mengfei Du,
  Xuancheng Ren, Rui Men, Dayiheng Liu, Chang Zhou, Jingren Zhou, and Junyang
  Lin.
\newblock {Qwen2-VL}: Enhancing vision-language model's perception of the world
  at any resolution, 2024{\natexlab{a}}.
\newblock URL \url{https://arxiv.org/abs/2409.12191}.

\bibitem[Bai et~al.(2025)]{bai2025qwen3vl}
Shuai Bai et~al.
\newblock {Qwen3-VL Technical Report}.
\newblock \emph{arXiv preprint arXiv:2511.21631}, 2025.
\newblock URL \url{https://arxiv.org/abs/2511.21631}.

\bibitem[{Foundation Model Team} and {XPeng Inc.}(2026)]{team2026xcot}
{Foundation Model Team} and {XPeng Inc.}
\newblock {XCoT-VLA: Executable Chain-of-Thought for Vision-Language-Action
  Driving}.
\newblock \emph{arXiv preprint arXiv:2608.10976}, 2026.

\bibitem[Tian et~al.(2024)Tian, Gu, Li, Liu, Wang, Zhao, Zhan, Jia, Lang, and
  Zhao]{tian2024drivevlm}
Xiaoyu Tian, Junru Gu, Bailin Li, Yicheng Liu, Yang Wang, Zhiyong Zhao, Kun
  Zhan, Peng Jia, Xianpeng Lang, and Hang Zhao.
\newblock {DriveVLM}: The convergence of autonomous driving and large
  vision-language models, 2024.
\newblock URL \url{https://arxiv.org/abs/2402.12289}.

\bibitem[Black et~al.(2024)Black, Brown, Driess, Esmail, Equi, Finn, Fusai,
  Groom, Hausman, Ichter, et~al.]{black2024pi0}
Kevin Black, Noah Brown, Danny Driess, Adnan Esmail, Michael Equi, Chelsea
  Finn, Niccolo Fusai, Lachy Groom, Karol Hausman, Brian Ichter, et~al.
\newblock $\pi_{0}$: A vision-language-action flow model for general robot
  control.
\newblock \emph{arXiv preprint arXiv:2410.24164}, 2024.
\newblock URL \url{https://arxiv.org/abs/2410.24164}.

\bibitem[Caesar et~al.(2020)Caesar, Bankiti, Lang, Vora, Liong, Xu, Krishnan,
  Pan, Baldan, and Beijbom]{caesar2020nuscenes}
Holger Caesar, Varun Bankiti, Alex~H. Lang, Sourabh Vora, Venice~Erin Liong,
  Qiang Xu, Anush Krishnan, Yu~Pan, Giancarlo Baldan, and Oscar Beijbom.
\newblock {nuScenes}: A multimodal dataset for autonomous driving, 2020.
\newblock URL \url{https://arxiv.org/abs/1903.11027}.

\bibitem[Khazatsky et~al.(2024)Khazatsky, Pertsch, Nair, Balakrishna, Dasari,
  Karamcheti, Nasiriany, Srirama, Chen, Ellis, et~al.]{khazatsky2024droid}
Alexander Khazatsky, Karl Pertsch, Suraj Nair, Ashwin Balakrishna, Sudeep
  Dasari, Siddharth Karamcheti, Soroush Nasiriany, Mohan~Kumar Srirama,
  Lawrence~Yunliang Chen, Kirsty Ellis, et~al.
\newblock {DROID: A Large-Scale In-The-Wild Robot Manipulation Dataset}.
\newblock \emph{arXiv preprint arXiv:2403.12945}, 2024.

\bibitem[Bolya et~al.(2023)Bolya, Fu, Dai, Zhang, Feichtenhofer, and
  Hoffman]{bolya2023tome}
Daniel Bolya, Cheng-Yang Fu, Xiaoliang Dai, Peizhao Zhang, Christoph
  Feichtenhofer, and Judy Hoffman.
\newblock Token merging: Your {ViT} but faster, 2023.
\newblock URL \url{https://arxiv.org/abs/2210.09461}.

\bibitem[Shang et~al.(2025)Shang, Cai, Xu, Lee, and Yan]{shang2025llava}
Yuzhang Shang, Mu~Cai, Bingxin Xu, Yong~Jae Lee, and Yan Yan.
\newblock {LLaVA-PruMerge}: Adaptive token reduction for efficient large
  multimodal models.
\newblock In \emph{2025 IEEE/CVF International Conference on Computer Vision
  (ICCV)}, pages 22857--22867. IEEE, 2025.

\bibitem[Huang et~al.(2025)Huang, Zhou, and Han]{huang2025prunevid}
Xiaohu Huang, Hao Zhou, and Kai Han.
\newblock {PruneVid}: Visual token pruning for efficient video large language
  models.
\newblock In \emph{Findings of the Association for Computational Linguistics:
  ACL 2025}, pages 19959--19973. Association for Computational Linguistics,
  2025.
\newblock \doi{10.18653/v1/2025.findings-acl.1024}.
\newblock URL \url{https://aclanthology.org/2025.findings-acl.1024/}.

\bibitem[Shen et~al.(2025)Shen, Gong, He, Zhang, Liu, Zhao, and
  Ding]{shen2026fastvid}
Leqi Shen, Guoqiang Gong, Tao He, Yifeng Zhang, Pengzhang Liu, Sicheng Zhao,
  and Guiguang Ding.
\newblock {FastVID}: Dynamic density pruning for fast video large language
  models.
\newblock \emph{Advances in Neural Information Processing Systems},
  38:\penalty0 123553--123581, 2025.
\newblock \doi{10.52202/085713-4118}.

\bibitem[Chen et~al.(2024{\natexlab{a}})Chen, Xu, Kirmani, Ichter, Driess,
  Florence, Sadigh, Guibas, and Xia]{chen2024spatialvlm}
Boyuan Chen, Zhuo Xu, Sean Kirmani, Brian Ichter, Danny Driess, Pete Florence,
  Dorsa Sadigh, Leonidas Guibas, and Fei Xia.
\newblock {SpatialVLM}: Endowing vision-language models with spatial reasoning
  capabilities, 2024{\natexlab{a}}.
\newblock URL \url{https://arxiv.org/abs/2401.12168}.

\bibitem[Ren et~al.(2024)Ren, Yao, Li, Sun, and Hou]{ren2024timechat}
Shuhuai Ren, Linli Yao, Shicheng Li, Xu~Sun, and Lu~Hou.
\newblock {TimeChat}: A time-sensitive multimodal large language model for long
  video understanding.
\newblock In \emph{Proceedings of the IEEE/CVF Conference on Computer Vision
  and Pattern Recognition (CVPR)}, pages 14313--14323, 2024.
\newblock \doi{10.1109/CVPR52733.2024.01357}.
\newblock URL
  \url{https://openaccess.thecvf.com/content/CVPR2024/html/Ren_TimeChat_A_Time-sensitive_Multimodal_Large_Language_Model_for_Long_Video_CVPR_2024_paper.html}.

\bibitem[{NVIDIA}(n.d.)]{nvidiaDriveAGXOrin}
{NVIDIA}.
\newblock {DRIVE AGX Developer Kits}.
\newblock \url{https://developer.nvidia.com/drive/agx/}, n.d.
\newblock DRIVE AGX Orin section. Accessed: 2026-11-05.

\bibitem[Cai et~al.(2026)Cai, Liu, Xiong, Liu, Chandra, and Shi]{cai2026vlm3}
Zhipeng Cai, Zhuang Liu, Yunyang Xiong, Zechun Liu, Vikas Chandra, and Yangyang
  Shi.
\newblock {VLM3}: Vision language models are native {3D} learners.
\newblock \emph{arXiv preprint arXiv:2605.30561}, 2026.
\newblock URL \url{https://arxiv.org/abs/2605.30561}.

\bibitem[Liu et~al.(2023{\natexlab{b}})Liu, Zhu, Gao, Feng, Liu, Zhu, and
  Stone]{liu2023libero}
Bo~Liu, Yifeng Zhu, Chongkai Gao, Yihao Feng, Qiang Liu, Yuke Zhu, and Peter
  Stone.
\newblock {LIBERO}: Benchmarking knowledge transfer for lifelong robot
  learning.
\newblock In \emph{Advances in Neural Information Processing Systems},
  volume~36, pages 44776--44791, 2023{\natexlab{b}}.

\bibitem[Faghri et~al.(2025)]{faghri2025mobileclip2}
Fartash Faghri et~al.
\newblock {MobileCLIP2: Improving Multi-Modal Reinforced Training}.
\newblock \emph{arXiv preprint arXiv:2508.20691}, 2025.
\newblock URL \url{https://arxiv.org/abs/2508.20691}.

\bibitem[Tschannen et~al.(2025)Tschannen, Gritsenko, Wang, Naeem,
  Alabdulmohsin, Parthasarathy, Evans, Beyer, Xia, Mustafa,
  et~al.]{tschannen2025siglip}
Michael Tschannen, Alexey Gritsenko, Xiao Wang, Muhammad~Ferjad Naeem, Ibrahim
  Alabdulmohsin, Nikhil Parthasarathy, Talfan Evans, Lucas Beyer, Ye~Xia, Basil
  Mustafa, et~al.
\newblock {SigLIP 2: Multilingual Vision-Language Encoders with Improved
  Semantic Understanding, Localization, and Dense Features}.
\newblock \emph{arXiv preprint arXiv:2502.14786}, 2025.

\bibitem[Wu et~al.(2026)Wu, Chen, et~al.]{wu2026turingvit}
Qiman Wu, Hanlin Chen, et~al.
\newblock {TuringViT: Making SOTA Vision Transformers Accessible to All}.
\newblock \emph{arXiv preprint arXiv:2606.24253}, 2026.
\newblock URL \url{https://arxiv.org/abs/2606.24253}.

\bibitem[Sim{\'e}oni et~al.(2025)Sim{\'e}oni, Vo, Seitzer, Baldassarre, Oquab,
  Jose, Khalidov, Szafraniec, Yi, Ramamonjisoa, et~al.]{simeoni2025dinov3}
Oriane Sim{\'e}oni, Huy~V. Vo, Maximilian Seitzer, Federico Baldassarre, Maxime
  Oquab, Cijo Jose, Vasil Khalidov, Marc Szafraniec, Seungeun Yi, Micha{\"e}l
  Ramamonjisoa, et~al.
\newblock {DINOv3}.
\newblock \emph{arXiv preprint arXiv:2508.10104}, 2025.

\bibitem[An et~al.(2026)An, Xie, Tang, Yan, Tan, Zhu, Chen, Zhao, Qin, Yang,
  Shen, Zhang, Zhang, Zhang, Cheng, Zhang, Wu, Ge, Ran, Song, Li, Feng, Hu,
  Chen, Niu, Li, Feng, Liu, Ge, and Deng]{an2026llavaonevision2}
Xiang An, Yin Xie, Feilong Tang, Yunyao Yan, Huajie Tan, Didi Zhu, Changrui
  Chen, Xiuwei Zhao, Bin Qin, Kaicheng Yang, Yifei Shen, Yuanhan Zhang, Kaichen
  Zhang, Wenkang Zhang, Zheng Cheng, Nansen Zhang, Chunsheng Wu, Chunjiang Ge,
  Zimin Ran, Dehua Song, Chunyuan Li, Shikun Feng, Ming Hu, Zhangquan Chen,
  Junbo Niu, Bo~Li, Ziyong Feng, Ziwei Liu, Zongyuan Ge, and Jiankang Deng.
\newblock {LLaVA-OneVision-2: Towards Next-Generation Perceptual Intelligence}.
\newblock \emph{arXiv preprint arXiv:2605.25979}, 2026.
\newblock URL \url{https://arxiv.org/abs/2605.25979}.

\bibitem[{Qwen} et~al.(2024){Qwen}, Yang, et~al.]{yang2024qwen25}
{Qwen}, An~Yang, et~al.
\newblock {Qwen2.5} technical report.
\newblock \emph{arXiv preprint arXiv:2412.15115}, 2024.
\newblock URL \url{https://arxiv.org/abs/2412.15115}.

\bibitem[Zhu et~al.(2025)]{zhu2025internvl3}
Jinguo Zhu et~al.
\newblock {InternVL3}: Exploring advanced training and test-time recipes for
  open-source multimodal models.
\newblock \emph{arXiv preprint arXiv:2504.10479}, 2025.
\newblock URL \url{https://arxiv.org/abs/2504.10479}.

\bibitem[Kazemzadeh et~al.(2014)Kazemzadeh, Ordonez, Matten, and
  Berg]{kazemzadeh2014referitgame}
Sahar Kazemzadeh, Vicente Ordonez, Mark Matten, and Tamara~L. Berg.
\newblock Referitgame: Referring to objects in photographs of natural scenes.
\newblock In \emph{Proceedings of the 2014 Conference on Empirical Methods in
  Natural Language Processing}, pages 787--798, 2014.

\bibitem[Yu et~al.(2016)Yu, Poirson, Yang, Berg, and Berg]{yu2016modeling}
Licheng Yu, Patrick Poirson, Shan Yang, Alexander~C. Berg, and Tamara~L. Berg.
\newblock Modeling context in referring expressions.
\newblock In \emph{Computer Vision -- ECCV 2016}, pages 69--85. Springer, 2016.

\bibitem[Paiss et~al.(2023)Paiss, Ephrat, Tov, Zada, Mosseri, Irani, and
  Dekel]{paiss2023teachingclip}
Roni Paiss, Ariel Ephrat, Omer Tov, Shiran Zada, Inbar Mosseri, Michal Irani,
  and Tali Dekel.
\newblock {Teaching CLIP to Count to Ten}.
\newblock In \emph{Proceedings of the IEEE/CVF International Conference on
  Computer Vision (ICCV)}, 2023.

\bibitem[Beyer et~al.(2024)Beyer, Steiner, {Susano Pinto},
  et~al.]{beyer2024paligemma}
Lucas Beyer, Andreas Steiner, Andr{\'e} {Susano Pinto}, et~al.
\newblock {PaliGemma}: A versatile {3B} {VLM} for transfer.
\newblock \emph{arXiv preprint arXiv:2407.07726}, 2024.
\newblock URL \url{https://arxiv.org/abs/2407.07726}.

\bibitem[Liu et~al.(2024)]{liu2024ocrbench}
Yuliang Liu et~al.
\newblock {OCRBench}: On the hidden mystery of {OCR} in large multimodal
  models.
\newblock \emph{Science China Information Sciences}, 67, 2024.

\bibitem[Fu et~al.(2025)Fu, Kuang, Song, Huang, Yang, Li, Zhu, Luo, Wang, Lu,
  et~al.]{fu2025ocrbenchv2}
Ling Fu, Zhebin Kuang, Jiajun Song, Mingxin Huang, Biao Yang, Yuzhe Li, Linghao
  Zhu, Qidi Luo, Xinyu Wang, Hao Lu, et~al.
\newblock {OCRBench v2: An Improved Benchmark for Evaluating Large Multimodal
  Models on Visual Text Localization and Reasoning}.
\newblock \emph{arXiv preprint arXiv:2501.00321}, 2025.

\bibitem[Masry et~al.(2022)Masry, Long, Tan, Joty, and Hoque]{masry2022chartqa}
Ahmed Masry, Do~Xuan Long, Jia~Qing Tan, Shafiq Joty, and Enamul Hoque.
\newblock {ChartQA}: A benchmark for question answering about charts with
  visual and logical reasoning.
\newblock In \emph{Findings of the Association for Computational Linguistics:
  ACL 2022}, pages 2263--2279, 2022.
\newblock \doi{10.18653/v1/2022.findings-acl.177}.

\bibitem[Singh et~al.(2019)Singh, Natarajan, Shah, Jiang, Chen, Batra, Parikh,
  and Rohrbach]{singh2019textvqa}
Amanpreet Singh, Vivek Natarajan, Meet Shah, Yu~Jiang, Xinlei Chen, Dhruv
  Batra, Devi Parikh, and Marcus Rohrbach.
\newblock Towards {VQA} models that can read.
\newblock In \emph{Proceedings of the IEEE/CVF International Conference on
  Computer Vision}, 2019.

\bibitem[Mathew et~al.(2021)Mathew, Karatzas, and Jawahar]{mathew2021docvqa}
Minesh Mathew, Dimosthenis Karatzas, and C.~V. Jawahar.
\newblock {DocVQA}: A dataset for {VQA} on document images.
\newblock In \emph{Proceedings of the IEEE/CVF Winter Conference on
  Applications of Computer Vision}, pages 2200--2209, 2021.

\bibitem[Kembhavi et~al.(2016)Kembhavi, Salvato, Kolve, Seo, Hajishirzi, and
  Farhadi]{kembhavi2016diagram}
Aniruddha Kembhavi, Mike Salvato, Eric Kolve, Minjoon Seo, Hannaneh Hajishirzi,
  and Ali Farhadi.
\newblock {A Diagram Is Worth A Dozen Images}.
\newblock In \emph{Computer Vision -- ECCV 2016}, pages 235--251, 2016.

\bibitem[Yue et~al.(2024)]{yue2024mmmu}
Xiang Yue et~al.
\newblock {MMMU}: A massive multi-discipline multimodal understanding and
  reasoning benchmark for expert {AGI}.
\newblock In \emph{Proceedings of the IEEE/CVF Conference on Computer Vision
  and Pattern Recognition}, 2024.

\bibitem[Wang et~al.(2024{\natexlab{b}})Wang, Pan, Shi, Lu, Zhan, and
  Li]{wang2024mathvision}
Ke~Wang, Junting Pan, Weikang Shi, Zimu Lu, Mingjie Zhan, and Hongsheng Li.
\newblock {Measuring Multimodal Mathematical Reasoning with MATH-Vision
  Dataset}.
\newblock \emph{arXiv preprint arXiv:2402.14804}, 2024{\natexlab{b}}.
\newblock URL \url{https://arxiv.org/abs/2402.14804}.

\bibitem[Lu et~al.(2024)Lu, Bansal, Xia, Liu, Li, Hajishirzi, Cheng, Chang,
  Galley, and Gao]{lu2024mathvista}
Pan Lu, Hritik Bansal, Tony Xia, Jiacheng Liu, Chunyuan Li, Hannaneh
  Hajishirzi, Hao Cheng, Kai-Wei Chang, Michel Galley, and Jianfeng Gao.
\newblock {MathVista: Evaluating Mathematical Reasoning of Foundation Models in
  Visual Contexts}.
\newblock In \emph{Proceedings of the International Conference on Learning
  Representations}, 2024.

\bibitem[Zou et~al.(2024)Zou, Guo, Yang, Zhang, Hu, and Zhang]{zou2024dynamath}
Chengke Zou, Xingang Guo, Rui Yang, Junyu Zhang, Bin Hu, and Huan Zhang.
\newblock {DynaMath: A Dynamic Visual Benchmark for Evaluating Mathematical
  Reasoning Robustness of Vision Language Models}.
\newblock \emph{arXiv preprint arXiv:2411.00836}, 2024.
\newblock URL \url{https://arxiv.org/abs/2411.00836}.

\bibitem[Xiao et~al.(2024)Xiao, Sun, Liu, and Wang]{xiao2024logicvista}
Yijia Xiao, Edward Sun, Tianyu Liu, and Wei Wang.
\newblock {LogicVista: Multimodal LLM Logical Reasoning Benchmark in Visual
  Contexts}.
\newblock \emph{arXiv preprint arXiv:2407.04973}, 2024.

\bibitem[Liu et~al.(2023{\natexlab{c}})]{liu2023mmbench}
Yuan Liu et~al.
\newblock {MMBench}: Is your multi-modal model an all-around player?
\newblock \emph{arXiv preprint arXiv:2307.06281}, 2023{\natexlab{c}}.

\bibitem[{xAI}(2024)]{xai2024realworldqa}
{xAI}.
\newblock {RealWorldQA} benchmark, 2024.
\newblock URL \url{https://x.ai/news/grok-1.5v}.
\newblock Dataset release.

\bibitem[Chen et~al.(2024{\natexlab{b}})]{chen2024mmstar}
Lin Chen et~al.
\newblock Are we on the right way for evaluating large vision-language models?
\newblock \emph{arXiv preprint arXiv:2403.20330}, 2024{\natexlab{b}}.

\bibitem[Yu et~al.(2023)Yu, Yang, Li, Wang, Lin, Liu, Wang, and
  Wang]{yu2023mmvet}
Weihao Yu, Zhengyuan Yang, Linjie Li, Jianfeng Wang, Kevin Lin, Zicheng Liu,
  Xinchao Wang, and Lijuan Wang.
\newblock {MM-Vet: Evaluating Large Multimodal Models for Integrated
  Capabilities}.
\newblock \emph{arXiv preprint arXiv:2308.02490}, 2023.

\bibitem[Guan et~al.(2024)]{guan2024hallusionbench}
Tianrui Guan et~al.
\newblock {HallusionBench}: An advanced diagnostic suite for entangled language
  hallucination and visual illusion in large vision-language models.
\newblock In \emph{Proceedings of the IEEE/CVF Conference on Computer Vision
  and Pattern Recognition}, 2024.

\bibitem[Zhang et~al.(2024)]{zhang2024mmerealworld}
Yi-Fan Zhang et~al.
\newblock {MME-RealWorld: Could Your Multimodal LLM Challenge High-Resolution
  Real-World Scenarios that are Difficult for Humans?}
\newblock \emph{arXiv preprint arXiv:2408.13257}, 2024.
\newblock URL \url{https://arxiv.org/abs/2408.13257}.

\bibitem[Li et~al.(2023)]{li2023seedbench}
Bohao Li et~al.
\newblock {SEED-Bench}: Benchmarking multimodal llms with generative
  comprehension.
\newblock \emph{arXiv preprint arXiv:2307.16125}, 2023.

\bibitem[Li et~al.(2024)]{li2024mvbench}
Kunchang Li et~al.
\newblock {MVBench}: A comprehensive multi-modal video understanding benchmark.
\newblock In \emph{Proceedings of the IEEE/CVF Conference on Computer Vision
  and Pattern Recognition}, 2024.

\bibitem[Zhou et~al.(2025)Zhou, Shu, Zhao, Wu, Liang, Xiao, Qin, Yang, Xiong,
  Zhang, et~al.]{zhou2025mlvu}
Junjie Zhou, Yan Shu, Bo~Zhao, Boya Wu, Zhengyang Liang, Shitao Xiao, Minghao
  Qin, Xi~Yang, Yongping Xiong, Bo~Zhang, et~al.
\newblock {MLVU: Benchmarking Multi-task Long Video Understanding}.
\newblock In \emph{Proceedings of the IEEE/CVF Conference on Computer Vision
  and Pattern Recognition}, 2025.

\bibitem[Wu et~al.(2024)]{wu2024longvideobench}
Haoning Wu et~al.
\newblock {LongVideoBench}: A benchmark for long-context interleaved
  video-language understanding.
\newblock \emph{arXiv preprint arXiv:2407.15754}, 2024.

\bibitem[Silberman et~al.(2012)Silberman, Hoiem, Kohli, and
  Fergus]{silberman2012indoor}
Nathan Silberman, Derek Hoiem, Pushmeet Kohli, and Rob Fergus.
\newblock Indoor segmentation and support inference from {RGBD} images.
\newblock In \emph{European Conference on Computer Vision (ECCV)}, pages
  746--760. Springer, 2012.

\bibitem[Sch{\"o}ps et~al.(2017)Sch{\"o}ps, Sch{\"o}nberger, Galliani, Sattler,
  Schindler, Pollefeys, and Geiger]{schops2017eth3d}
Thomas Sch{\"o}ps, Johannes~L. Sch{\"o}nberger, Silvano Galliani, Torsten
  Sattler, Konrad Schindler, Marc Pollefeys, and Andreas Geiger.
\newblock A multi-view stereo benchmark with high-resolution images and
  multi-camera videos.
\newblock In \emph{Proceedings of the IEEE Conference on Computer Vision and
  Pattern Recognition (CVPR)}, pages 3260--3269, 2017.

\bibitem[Shotton et~al.(2013)Shotton, Glocker, Zach, Izadi, Criminisi, and
  Fitzgibbon]{shotton2013scene}
Jamie Shotton, Ben Glocker, Christopher Zach, Shahram Izadi, Antonio Criminisi,
  and Andrew Fitzgibbon.
\newblock Scene coordinate regression forests for camera relocalization in
  {RGB-D} images.
\newblock In \emph{Proceedings of the IEEE Conference on Computer Vision and
  Pattern Recognition (CVPR)}, pages 2930--2937, 2013.

\bibitem[Lin et~al.(2025)Lin, Chen, Liew, Chen, Li, Shi, Feng, and
  Kang]{lin2025depthanything3}
Haotong Lin, Sili Chen, Junhao Liew, Donny~Y. Chen, Zhenyu Li, Guang Shi,
  Jiashi Feng, and Bingyi Kang.
\newblock {Depth Anything 3: Recovering the Visual Space from Any Views}.
\newblock \emph{arXiv preprint arXiv:2511.10647}, 2025.
\newblock URL \url{https://arxiv.org/abs/2511.10647}.

\bibitem[Brazil et~al.(2023)Brazil, Kumar, Straub, Ravi, Johnson, and
  Gkioxari]{brazil2023omni3d}
Garrick Brazil, Abhinav Kumar, Julian Straub, Nikhila Ravi, Justin Johnson, and
  Georgia Gkioxari.
\newblock {Omni3D}: A large benchmark and model for {3D} object detection in
  the wild.
\newblock In \emph{Proceedings of the IEEE/CVF Conference on Computer Vision
  and Pattern Recognition (CVPR)}, pages 13154--13164, 2023.

\bibitem[Baruch et~al.(2021)Baruch, Chen, Dehghan, Dimry, Feigin, Fu, Gebauer,
  Joffe, Kurz, Schwartz, and Shulman]{baruch2021arkitscenes}
Gilad Baruch, Zhuoyuan Chen, Afshin Dehghan, Tal Dimry, Yuri Feigin, Peter Fu,
  Thomas Gebauer, Brandon Joffe, Daniel Kurz, Arik Schwartz, and Elad Shulman.
\newblock {ARKitScenes}: A diverse real-world dataset for {3D} indoor scene
  understanding using mobile {RGB-D} data.
\newblock In \emph{Advances in Neural Information Processing Systems: Datasets
  and Benchmarks}, 2021.

\bibitem[Dai et~al.(2017)Dai, Chang, Savva, Halber, Funkhouser, and
  Nie{\ss}ner]{dai2017scannet}
Angela Dai, Angel~X. Chang, Manolis Savva, Maciej Halber, Thomas Funkhouser,
  and Matthias Nie{\ss}ner.
\newblock {ScanNet}: Richly-annotated {3D} reconstructions of indoor scenes.
\newblock In \emph{Proceedings of the IEEE Conference on Computer Vision and
  Pattern Recognition (CVPR)}, 2017.

\bibitem[Song et~al.(2015)Song, Lichtenberg, and Xiao]{song2015sunrgbd}
Shuran Song, Samuel~P. Lichtenberg, and Jianxiong Xiao.
\newblock {SUN RGB-D}: A {RGB-D} scene understanding benchmark suite.
\newblock In \emph{Proceedings of the IEEE Conference on Computer Vision and
  Pattern Recognition (CVPR)}, 2015.

\bibitem[Fu et~al.(2024)]{fu2024blink}
Xingyu Fu et~al.
\newblock {BLINK}: Multimodal large language models can see but not perceive.
\newblock In \emph{European Conference on Computer Vision}, 2024.

\bibitem[Du et~al.(2024)Du, Wu, Li, Huang, and Wei]{du2024embspatial}
Mengfei Du, Binhao Wu, Zejun Li, Xuanjing Huang, and Zhongyu Wei.
\newblock {EmbSpatial-Bench}: Benchmarking spatial understanding for embodied
  tasks with large vision-language models.
\newblock \emph{arXiv preprint arXiv:2406.05756}, 2024.

\bibitem[{Gemini Robotics Team}(2025)]{gemini2025robotics}
{Gemini Robotics Team}.
\newblock Gemini robotics: Bringing {AI} into the physical world.
\newblock \emph{arXiv preprint arXiv:2503.20020}, 2025.

\bibitem[Song et~al.(2025)Song, Blukis, Tremblay, Tyree, Su, and
  Birchfield]{song2025robospatial}
Chan~Hee Song, Valts Blukis, Jonathan Tremblay, Stephen Tyree, Yu~Su, and Stan
  Birchfield.
\newblock {RoboSpatial}: Teaching spatial understanding to {2D} and {3D}
  vision-language models for robotics.
\newblock In \emph{Proceedings of the IEEE/CVF Conference on Computer Vision
  and Pattern Recognition (CVPR)}, pages 15768--15780, 2025.

\bibitem[Yang et~al.(2025)Yang, Yang, Gupta, Han, Fei-Fei, and
  Xie]{yang2025vsibench}
Jihan Yang, Shusheng Yang, Anjali~W. Gupta, Rilyn Han, Li~Fei-Fei, and Saining
  Xie.
\newblock Thinking in space: How multimodal large language models see,
  remember, and recall spaces.
\newblock In \emph{Proceedings of the IEEE/CVF Conference on Computer Vision
  and Pattern Recognition (CVPR)}, pages 10632--10643, 2025.

\bibitem[Deng et~al.(2025)Deng, Pei, Zhang, and Du]{deng2025attribution}
Huiqi Deng, Hongbin Pei, Quanshi Zhang, and Mengnan Du.
\newblock Attribution explanations for deep neural networks: A theoretical
  perspective.
\newblock \emph{arXiv preprint arXiv:2508.07636}, 2025.
\newblock URL \url{https://arxiv.org/abs/2508.07636}.

\bibitem[Itti and Koch(2001)]{itti2001computational}
Laurent Itti and Christof Koch.
\newblock Computational modelling of visual attention.
\newblock \emph{Nature Reviews Neuroscience}, 2\penalty0 (3):\penalty0
  194--203, 2001.
\newblock \doi{10.1038/35058500}.

\bibitem[Simonyan et~al.(2013)Simonyan, Vedaldi, and
  Zisserman]{simonyan2013deep}
Karen Simonyan, Andrea Vedaldi, and Andrew Zisserman.
\newblock Deep inside convolutional networks: Visualising image classification
  models and saliency maps.
\newblock \emph{arXiv preprint arXiv:1312.6034}, 2013.
\newblock URL \url{https://arxiv.org/abs/1312.6034}.

\bibitem[Adebayo et~al.(2018)Adebayo, Gilmer, Muelly, Goodfellow, Hardt, and
  Kim]{adebayo2018sanity}
Julius Adebayo, Justin Gilmer, Michael Muelly, Ian Goodfellow, Moritz Hardt,
  and Been Kim.
\newblock Sanity checks for saliency maps.
\newblock In \emph{Advances in Neural Information Processing Systems},
  volume~31, 2018.

\bibitem[Jiang et~al.(2025)Jiang, Huang, Qian, Luo, Zhu, Zhong, Tang, Kong,
  Wang, Jiao, et~al.]{jiang2025vla4ad}
Sicong Jiang, Zilin Huang, Kangan Qian, Ziang Luo, Tianze Zhu, Yang Zhong,
  Yihong Tang, Menglin Kong, Yunlong Wang, Siwen Jiao, et~al.
\newblock A survey on vision-language-action models for autonomous driving.
\newblock \emph{arXiv preprint arXiv:2506.24044}, 2025.
\newblock URL \url{https://arxiv.org/abs/2506.24044}.

\bibitem[Sundararajan et~al.(2017)Sundararajan, Taly, and
  Yan]{sundararajan2017integrated}
Mukund Sundararajan, Ankur Taly, and Qiqi Yan.
\newblock Axiomatic attribution for deep networks.
\newblock In \emph{Proceedings of the International Conference on Machine
  Learning (ICML)}, 2017.

\bibitem[{Physical Intelligence} et~al.(2025){Physical Intelligence}, Black,
  Brown, Darpinian, Dhabalia, Driess, Esmail, Equi, Finn, Fusai,
  et~al.]{black2025pi05}
{Physical Intelligence}, Kevin Black, Noah Brown, James Darpinian, Karan
  Dhabalia, Danny Driess, Adnan Esmail, Michael Equi, Chelsea Finn, Niccolo
  Fusai, et~al.
\newblock $\pi_{0.5}$: A vision-language-action model with open-world
  generalization.
\newblock \emph{arXiv preprint arXiv:2504.16054}, 2025.
\newblock URL \url{https://arxiv.org/abs/2504.16054}.

\bibitem[{Physical Intelligence}(n.d.)]{openpiLiberoResults}
{Physical Intelligence}.
\newblock {OpenPI LIBERO Benchmark Results}, n.d.
\newblock URL
  \url{https://github.com/Physical-Intelligence/openpi/tree/main/examples/libero#results}.
\newblock Accessed 2026-11-18.

\bibitem[Yang et~al.(2026)Yang, Wang, Wang, Guo, Yu, Liu, Xu, Dong, and
  Li]{yang2026eaglevla}
Zebin Yang, Qi~Wang, Yunhe Wang, Xiurui Guo, Bo~Yu, Shaoshan Liu, Jiafeng Xu,
  Hao Dong, and Meng Li.
\newblock {EagleVLA: Towards Onboard Real-Time Robot Control via
  Foresight-Aligned Asynchronous Inference}.
\newblock \emph{arXiv preprint arXiv:2607.12659}, 2026.
\newblock URL \url{https://arxiv.org/abs/2607.12659}.

\bibitem[Tang et~al.(2025)Tang, Sun, Zhao, Yang, Lin, Zhang, Hou, Lu, Liu, and
  Han]{tang2025vlash}
Jiaming Tang, Yufei Sun, Yilong Zhao, Shang Yang, Yujun Lin, Zhuoyang Zhang,
  James Hou, Yao Lu, Zhijian Liu, and Song Han.
\newblock {VLASH: Real-Time VLAs via Future-State-Aware Asynchronous
  Inference}.
\newblock \emph{arXiv preprint arXiv:2512.01031}, 2025.
\newblock URL \url{https://arxiv.org/abs/2512.01031}.

\end{thebibliography}
\endgroup

\clearpage
\appendix
\section*{Appendix}

\raggedbottom
\section{Spatial Supervision Case Figures}
\label{app:spatial-supervision-cases}

\begin{figure}[!htbp]
  \centering
  \includegraphics[width=0.9\linewidth,height=0.42\textheight,keepaspectratio]{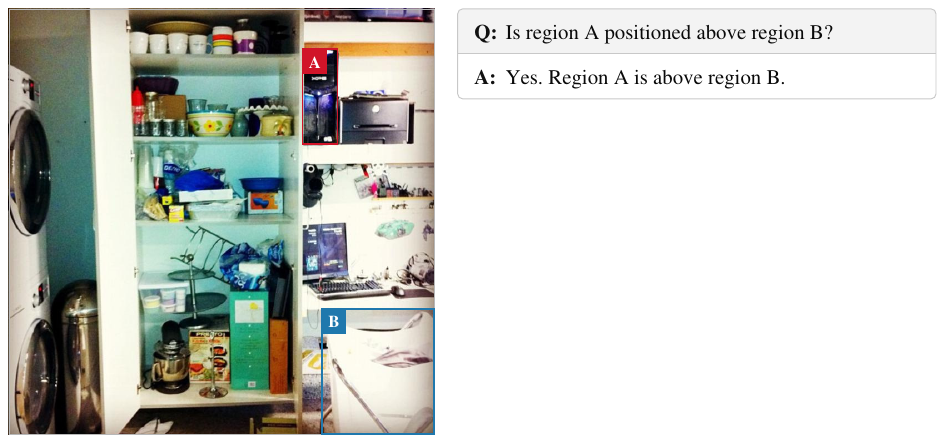}
  \caption{\textbf{Box-grounded spatial-relation example.}}
  \label{fig:data-case-spatial-relations}
\end{figure}

\begin{figure}[!htbp]
  \centering
  \includegraphics[width=0.9\linewidth,height=0.42\textheight,keepaspectratio]{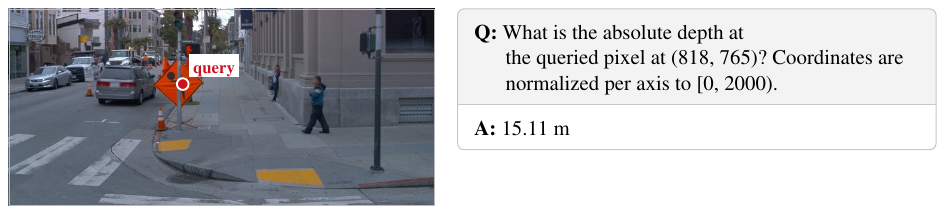}
  \caption{\textbf{Metric-depth estimation example.}}
  \label{fig:data-case-depth}
\end{figure}

\begin{figure}[!htbp]
  \centering
  \includegraphics[width=0.9\linewidth,height=0.42\textheight,keepaspectratio]{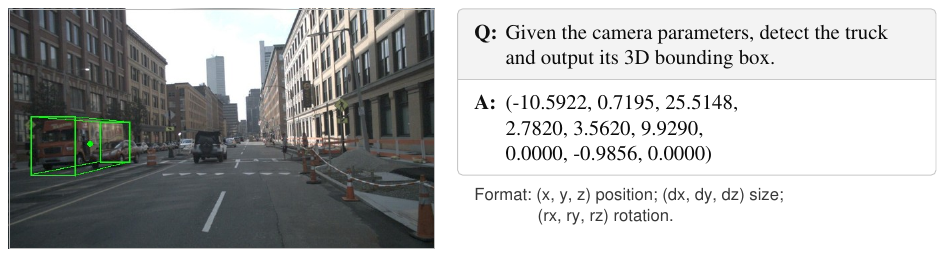}
  \caption{\textbf{Camera-conditioned 3D object-localization example.}}
  \label{fig:data-case-3d-geometry}
\end{figure}

\begin{figure}[!htbp]
  \centering
  \includegraphics[width=0.9\linewidth,height=0.42\textheight,keepaspectratio]{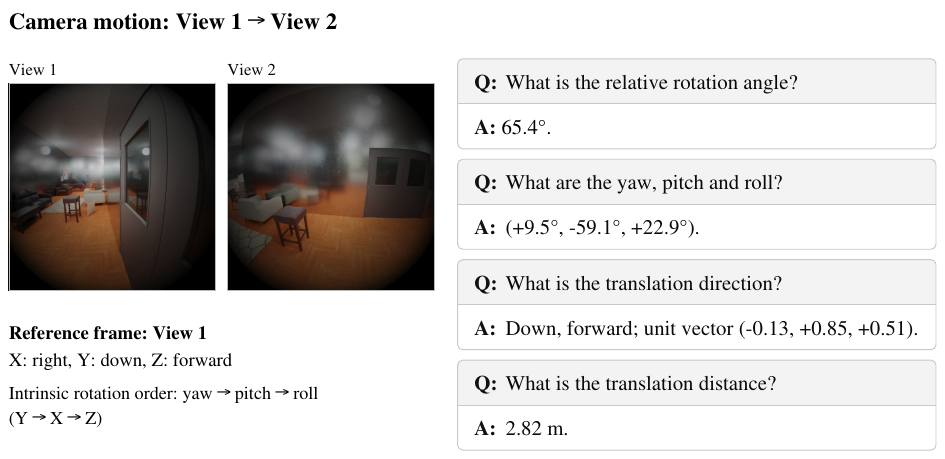}
  \caption{\textbf{Relative camera-pose example with rotation and translation questions.}}
  \label{fig:data-case-camera-pose}
\end{figure}

\begin{figure}[!htbp]
  \centering
  \includegraphics[width=0.9\linewidth,height=0.42\textheight,keepaspectratio]{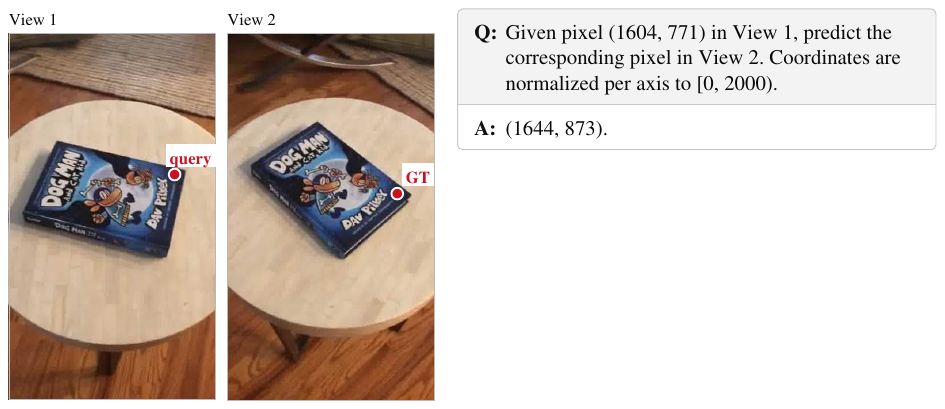}
  \caption{\textbf{Cross-view pixel-correspondence example.} The query pixel and its labeled match are shown in separate views.}
  \label{fig:data-case-pixel-correspondence}
\end{figure}

\begin{figure}[!htbp]
  \centering
  \includegraphics[width=0.9\linewidth,height=0.42\textheight,keepaspectratio]{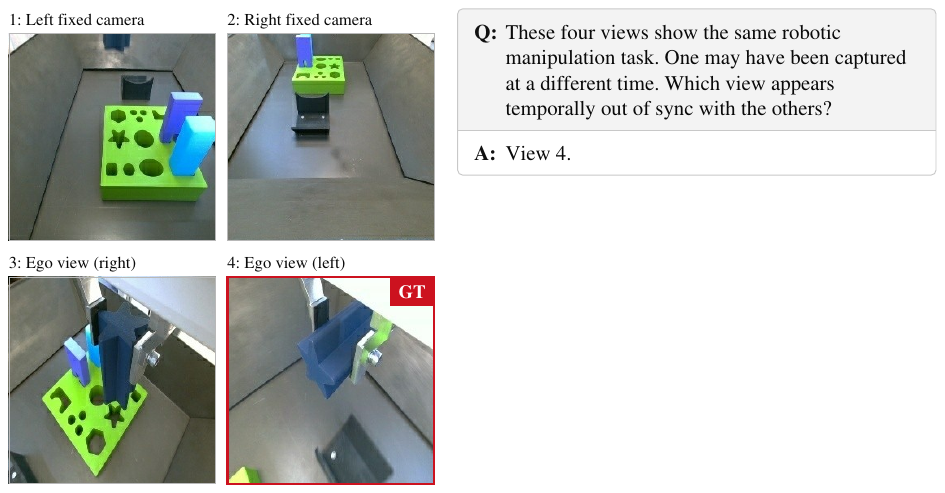}
  \caption{\textbf{Cross-view synchronization example for a robotic-manipulation scene.} Numbered images identify camera views; the highlighted view indicates the answer.}
  \label{fig:data-case-cross-view-sync}
\end{figure}

\begin{figure}[!htbp]
  \centering
  \includegraphics[width=0.9\linewidth,height=0.42\textheight,keepaspectratio]{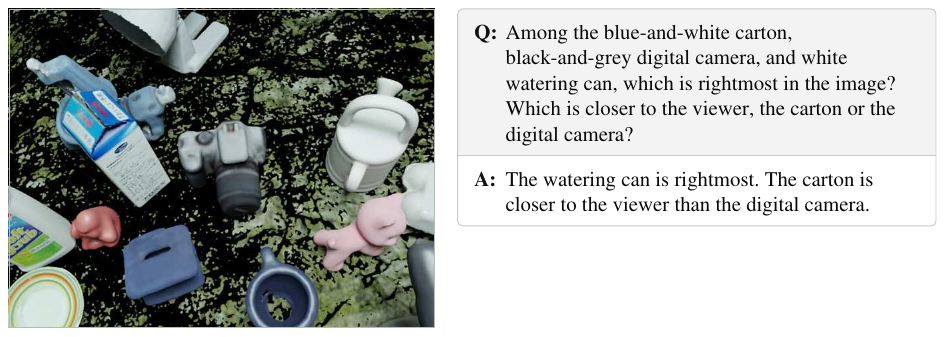}
  \caption{\textbf{Tabletop embodied spatial-reasoning example involving object identification, relative position, and proximity.}}
  \label{fig:data-case-embodied-spatial}
\end{figure}

\begin{figure}[!htbp]
  \centering
  \includegraphics[width=0.9\linewidth,height=0.42\textheight,keepaspectratio]{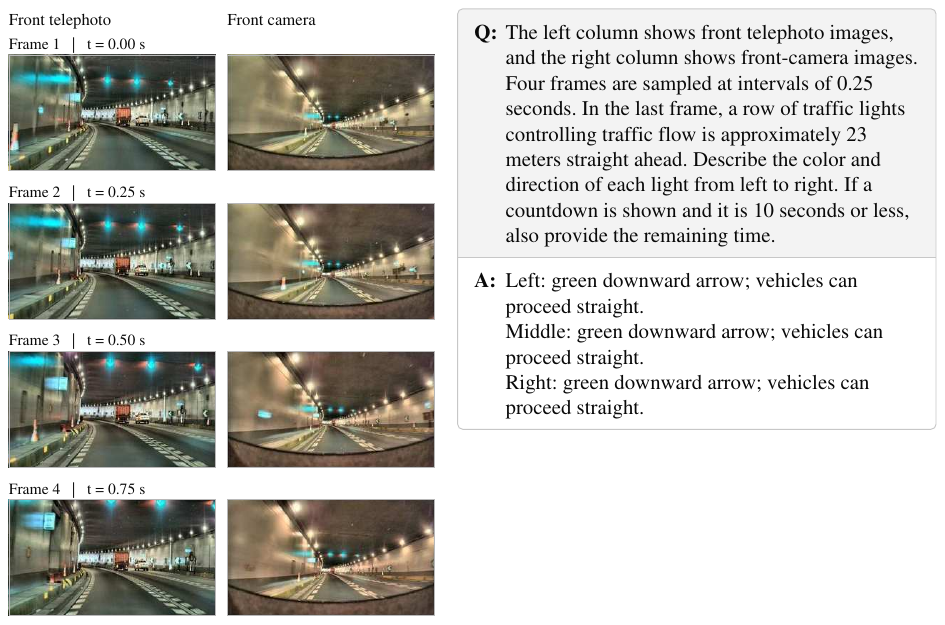}
  \caption{\textbf{Example from XPeng's internal alignment data for traffic-light-state understanding.} The left and right columns show front-telephoto and front-camera views, respectively, over four frames sampled at 0.25\,s intervals. The prompt asks for each signal's color and direction and any countdown of 10\,s or less.}
  \label{fig:data-case-xpeng-traffic-light-state}
\end{figure}

\begin{figure}[!htbp]
  \centering
  \includegraphics[width=0.9\linewidth,height=0.42\textheight,keepaspectratio]{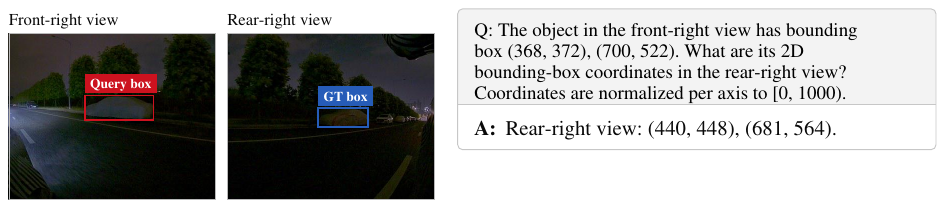}
  \caption{\textbf{Example from XPeng's internal alignment data for cross-view box grounding.} Given an object's 2D bounding box in the front-right view, the task predicts its corresponding box in the rear-right view.}
  \label{fig:data-case-xpeng-cross-view-box}
\end{figure}

\clearpage
\raggedbottom

\section{Per-Metric CCES Breakdown}
\label{app:cces-breakdown}

\begin{table}[htbp]
  \caption{\textbf{Per-metric CCES breakdown of HybridViT-MLQ against
  HybridViT (1.00).} Each Score is the ratio of an unnormalized cost to
  that of HybridViT; values below 1.00 indicate lower costs.
  $\Delta$Score (\%) is computed from displayed Score values as
  $100(\mathrm{Score}-1)$. Class averages are in italics.}
  \label{app:cces-full}
  \centering
  \small
  \setlength{\tabcolsep}{7pt}
  \renewcommand{\arraystretch}{1.18}
  \begin{tabular}{@{}llcc@{}}
    \toprule
    Category & Subclass & Score & $\Delta$Score (\%) \\
    \midrule
    \multirow{7}{*}{Compliance}
      & Off-center driving        & 1.009  & $+0.9$ \\
      & Oncoming-lane entry       & 1.010  & $+1.0$ \\
      & Traffic-light issues      & 0.997  & $-0.3$ \\
      & Waiting-area adherence    & 0.992  & $-0.8$ \\
      & Restricted-lane entry    & 0.999  & $-0.1$ \\
      & Line crossing            & 1.004  & $+0.4$ \\
      & \textit{Class average}    & \textit{1.002}  & \textit{$+0.2$} \\
    \midrule
    \multirow{4}{*}{Comfort}
      & Acceleration jerk         & 1.061  & $+6.1$ \\
      & Swaying                   & 0.977  & $-2.3$ \\
      & Abnormal decel.           & 0.976  & $-2.4$ \\
      & \textit{Class average}    & \textit{1.004}  & \textit{$+0.4$} \\
    \midrule
    \multirow{7}{*}{Efficiency}
      & Slow acceleration         & 0.992  & $-0.8$ \\
      & Speed-limit shortfall     & 0.973  & $-2.7$ \\
      & Delayed lane change       & 0.996  & $-0.4$ \\
      & Irregular detour          & 0.984  & $-1.6$ \\
      & Long following dist.      & 1.003  & $+0.3$ \\
      & Long stopping dist.       & 1.002  & $+0.2$ \\
      & \textit{Class average}    & \textit{0.992}  & \textit{$-0.8$} \\
    \midrule
    \multirow{7}{*}{Safety}
      & Insufficient braking      & 0.998  & $-0.2$ \\
      & Roadside collision        & 1.015  & $+1.5$ \\
      & Oncoming avoidance        & 0.995  & $-0.5$ \\
      & Risky lane change         & 0.997  & $-0.3$ \\
      & Close following           & 1.002  & $+0.2$ \\
      & Rear collision risk       & 0.962  & $-3.8$ \\
      & \textit{Class average}    & \textit{0.995}  & \textit{$-0.5$} \\
    \midrule
    \multirow{6}{*}{Navigation}
      & Missed/late lane change   & 0.938  & $-6.2$ \\
      & Main/side-road merge      & 1.008  & $+0.8$ \\
      & Intersection              & 0.972  & $-2.8$ \\
      & No-gap lane changes       & 1.111  & $+11.1$ \\
      & Wrong route               & 0.934  & $-6.6$ \\
      & \textit{Class average}    & \textit{0.991}  & \textit{$-0.9$} \\
    \bottomrule
  \end{tabular}
\end{table}

\clearpage
\section{Further Attribution Cases}
\label{app:saliency-cases}

\newcommand{\figSevenCamHeader}[2]{%
  \begin{minipage}[b]{#1}\centering\scriptsize #2\end{minipage}%
}
\newcommand{\figSevenCamImage}[2]{%
  \begin{minipage}[c]{#1}\centering#2\end{minipage}%
}
\newcommand{\figSevenCamRowLabel}[1]{%
  \begin{minipage}[c][1.17cm][c]{0.5cm}\centering\rotatebox{90}{\fontsize{4.75}{5.7}\selectfont #1}\end{minipage}%
}
\newcommand{\figSevenCamRow}[8]{%
  \par\noindent\makebox[\linewidth][c]{%
    \figSevenCamRowLabel{#1}%
    \figSevenCamImage{1.755cm}{#2}%
    \figSevenCamImage{1.560cm}{#3}%
    \figSevenCamImage{2.048cm}{#4}%
    \figSevenCamImage{2.340cm}{#5}%
    \figSevenCamImage{2.340cm}{#6}%
    \figSevenCamImage{1.560cm}{#7}%
    \figSevenCamImage{1.755cm}{#8}%
  }\par\vspace{0.5mm}%
}

\begin{figure}[H]
  \centering
  \makebox[\linewidth][c]{%
    \makebox[0.5cm]{}%
    \figSevenCamHeader{1.755cm}{Left rear}%
    \figSevenCamHeader{1.560cm}{Left front}%
    \figSevenCamHeader{2.048cm}{Front narrow}%
    \figSevenCamHeader{2.340cm}{Front fisheye}%
    \figSevenCamHeader{2.340cm}{Rear}%
    \figSevenCamHeader{1.560cm}{Right front}%
    \figSevenCamHeader{1.755cm}{Right rear}%
  }\par\vspace{0.8mm}
  \figSevenCamRow
    {Camera}%
    {\includegraphics[height=1.17cm]{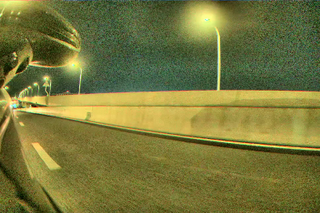}}%
    {\includegraphics[height=1.17cm]{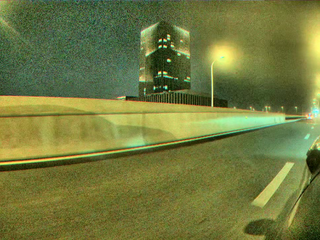}}%
    {\includegraphics[height=1.17cm]{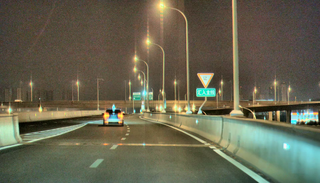}}%
    {\includegraphics[height=1.17cm]{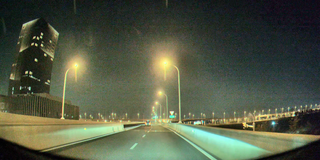}}%
    {\includegraphics[height=1.17cm]{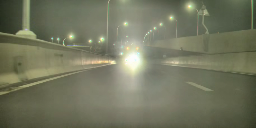}}%
    {\includegraphics[height=1.17cm]{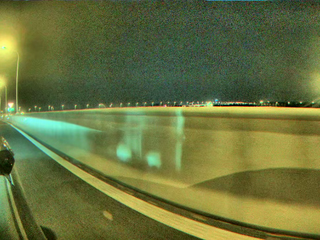}}%
    {\includegraphics[height=1.17cm]{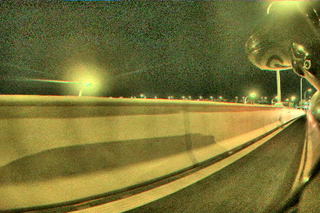}}%
  \figSevenCamRow
    {HybridViT}%
    {\includegraphics[height=1.17cm]{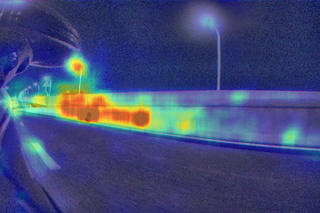}}%
    {\includegraphics[height=1.17cm]{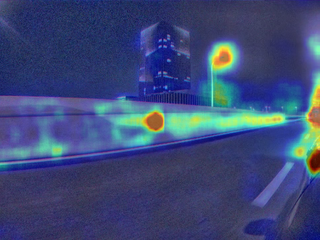}}%
    {\includegraphics[height=1.17cm]{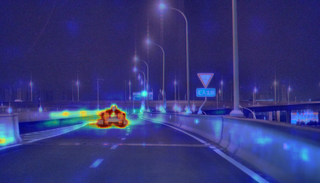}}%
    {\includegraphics[height=1.17cm]{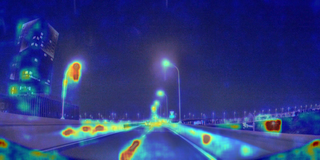}}%
    {\includegraphics[height=1.17cm]{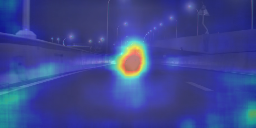}}%
    {\includegraphics[height=1.17cm]{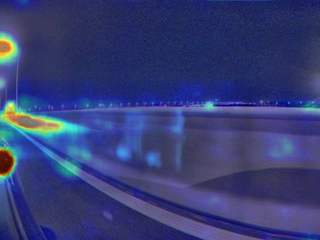}}%
    {\includegraphics[height=1.17cm]{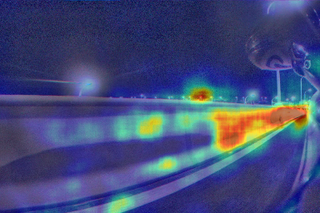}}%
  \figSevenCamRow
    {HybridViT-MLQ}%
    {\includegraphics[height=1.17cm]{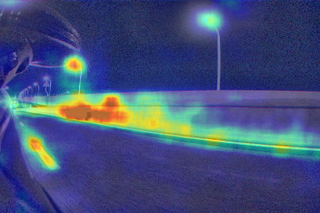}}%
    {\includegraphics[height=1.17cm]{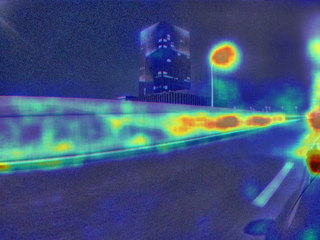}}%
    {\includegraphics[height=1.17cm]{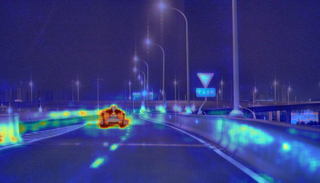}}%
    {\includegraphics[height=1.17cm]{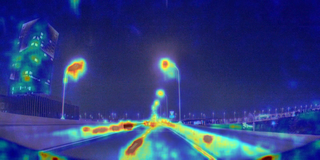}}%
    {\includegraphics[height=1.17cm]{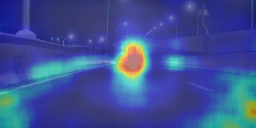}}%
    {\includegraphics[height=1.17cm]{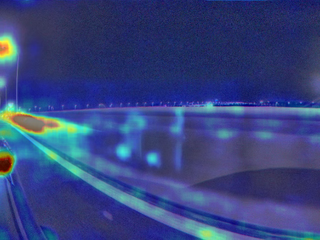}}%
    {\includegraphics[height=1.17cm]{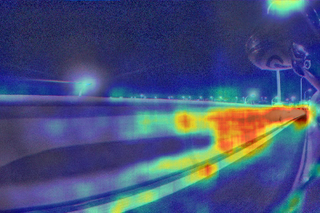}}%
  \vspace{1.2mm}

  \figSevenCamRow
    {Camera}%
    {\includegraphics[height=1.17cm]{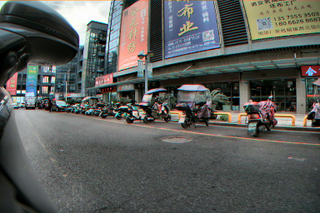}}%
    {\includegraphics[height=1.17cm]{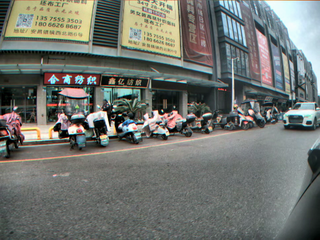}}%
    {\includegraphics[height=1.17cm]{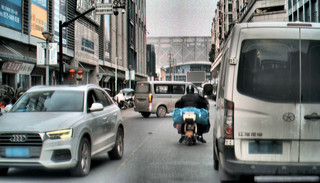}}%
    {\includegraphics[height=1.17cm]{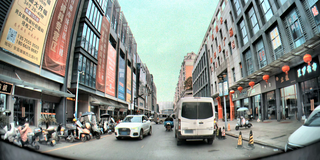}}%
    {\includegraphics[height=1.17cm]{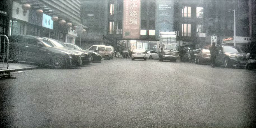}}%
    {\includegraphics[height=1.17cm]{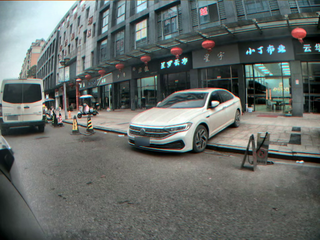}}%
    {\includegraphics[height=1.17cm]{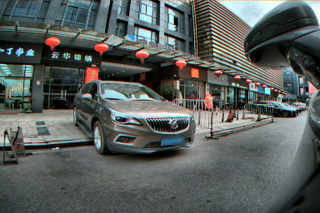}}%
  \figSevenCamRow
    {HybridViT}%
    {\includegraphics[height=1.17cm]{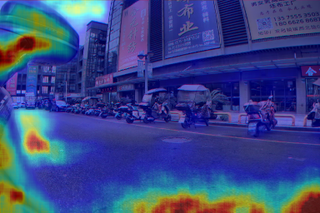}}%
    {\includegraphics[height=1.17cm]{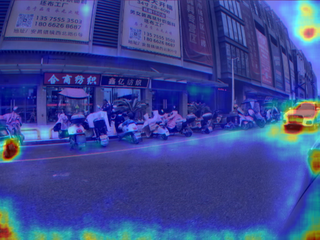}}%
    {\includegraphics[height=1.17cm]{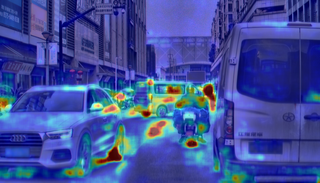}}%
    {\includegraphics[height=1.17cm]{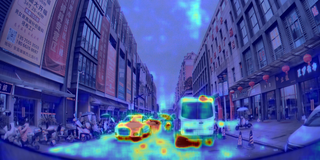}}%
    {\includegraphics[height=1.17cm]{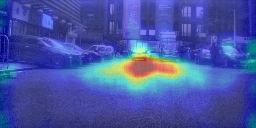}}%
    {\includegraphics[height=1.17cm]{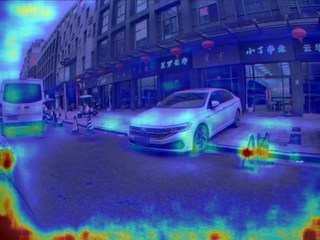}}%
    {\includegraphics[height=1.17cm]{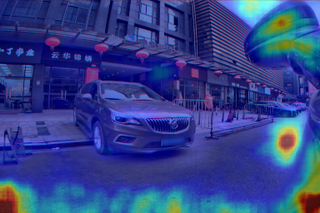}}%
  \figSevenCamRow
    {HybridViT-MLQ}%
    {\includegraphics[height=1.17cm]{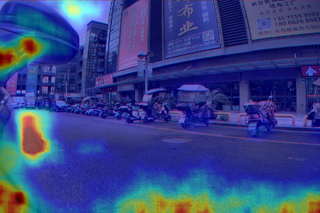}}%
    {\includegraphics[height=1.17cm]{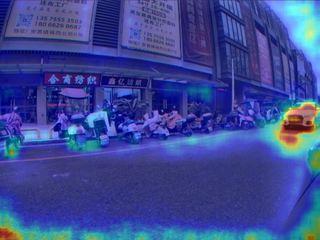}}%
    {\includegraphics[height=1.17cm]{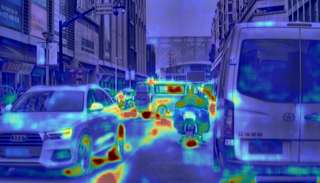}}%
    {\includegraphics[height=1.17cm]{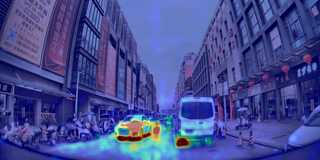}}%
    {\includegraphics[height=1.17cm]{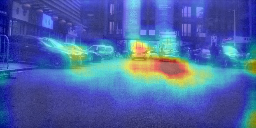}}%
    {\includegraphics[height=1.17cm]{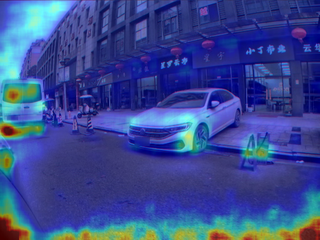}}%
    {\includegraphics[height=1.17cm]{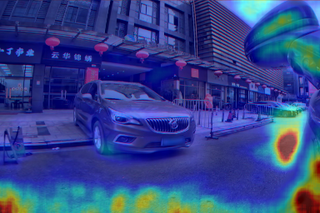}}%
  \vspace{1.2mm}

  \figSevenCamRow
    {Camera}%
    {\includegraphics[height=1.17cm]{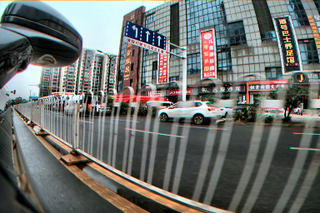}}%
    {\includegraphics[height=1.17cm]{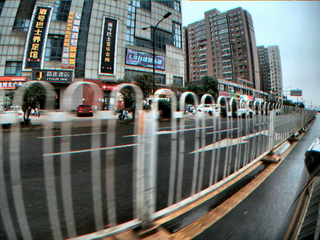}}%
    {\includegraphics[height=1.17cm]{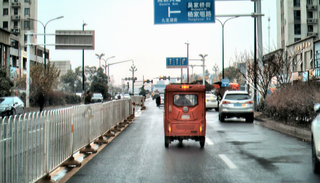}}%
    {\includegraphics[height=1.17cm]{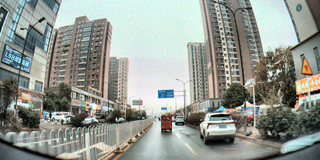}}%
    {\includegraphics[height=1.17cm]{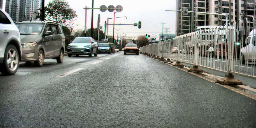}}%
    {\includegraphics[height=1.17cm]{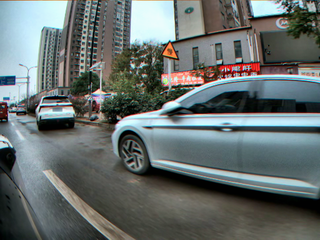}}%
    {\includegraphics[height=1.17cm]{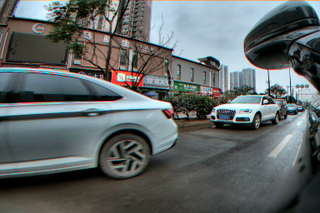}}%
  \figSevenCamRow
    {HybridViT}%
    {\includegraphics[height=1.17cm]{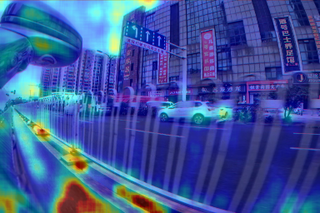}}%
    {\includegraphics[height=1.17cm]{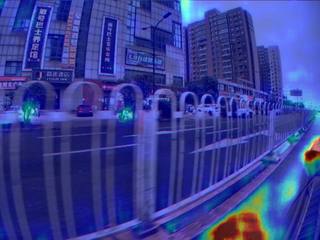}}%
    {\includegraphics[height=1.17cm]{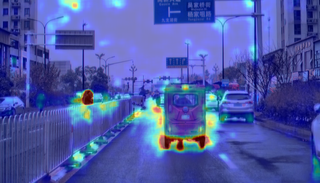}}%
    {\includegraphics[height=1.17cm]{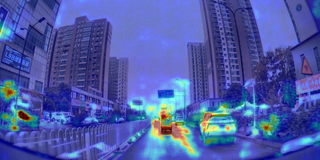}}%
    {\includegraphics[height=1.17cm]{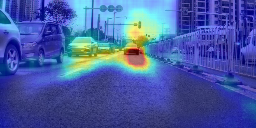}}%
    {\includegraphics[height=1.17cm]{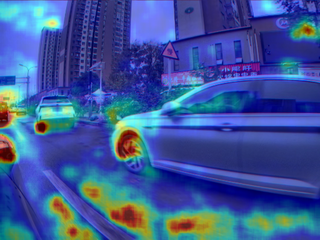}}%
    {\includegraphics[height=1.17cm]{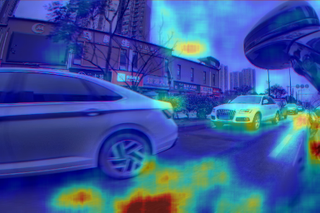}}%
  \figSevenCamRow
    {HybridViT-MLQ}%
    {\includegraphics[height=1.17cm]{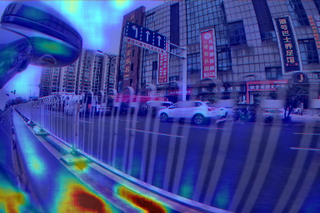}}%
    {\includegraphics[height=1.17cm]{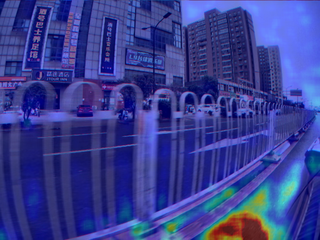}}%
    {\includegraphics[height=1.17cm]{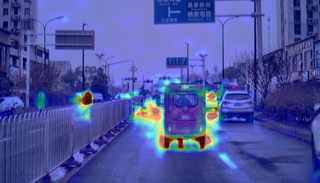}}%
    {\includegraphics[height=1.17cm]{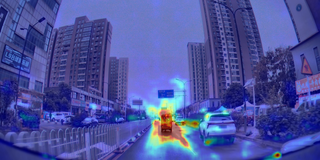}}%
    {\includegraphics[height=1.17cm]{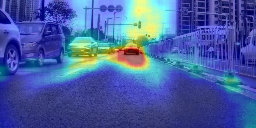}}%
    {\includegraphics[height=1.17cm]{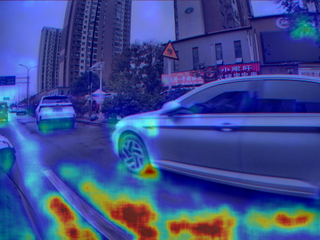}}%
    {\includegraphics[height=1.17cm]{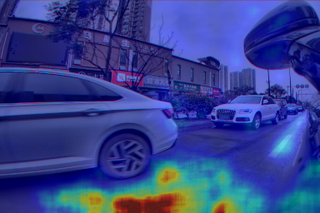}}%
  \vspace{1.2mm}

  \figSevenCamRow
    {Camera}%
    {\includegraphics[height=1.17cm]{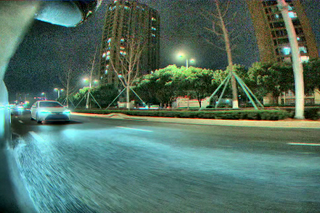}}%
    {\includegraphics[height=1.17cm]{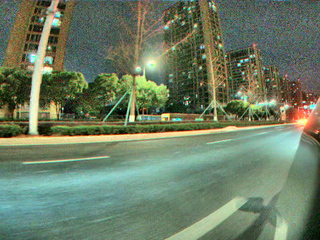}}%
    {\includegraphics[height=1.17cm]{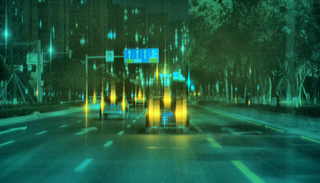}}%
    {\includegraphics[height=1.17cm]{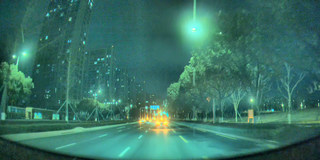}}%
    {\includegraphics[height=1.17cm]{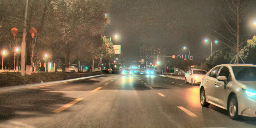}}%
    {\includegraphics[height=1.17cm]{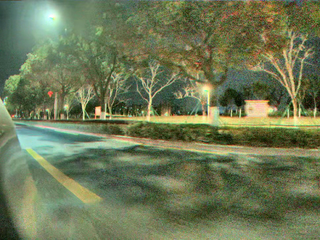}}%
    {\includegraphics[height=1.17cm]{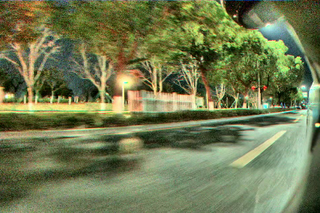}}%
  \figSevenCamRow
    {HybridViT}%
    {\includegraphics[height=1.17cm]{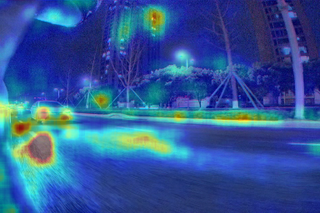}}%
    {\includegraphics[height=1.17cm]{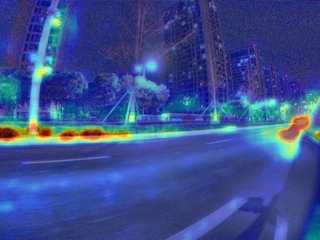}}%
    {\includegraphics[height=1.17cm]{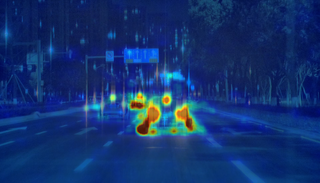}}%
    {\includegraphics[height=1.17cm]{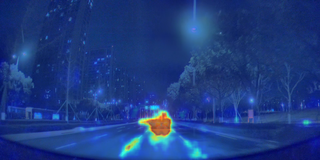}}%
    {\includegraphics[height=1.17cm]{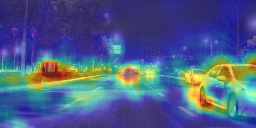}}%
    {\includegraphics[height=1.17cm]{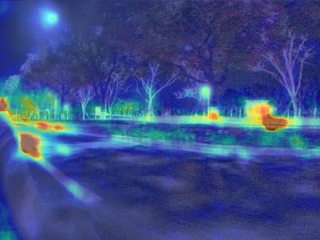}}%
    {\includegraphics[height=1.17cm]{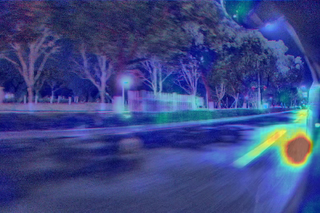}}%
  \figSevenCamRow
    {HybridViT-MLQ}%
    {\includegraphics[height=1.17cm]{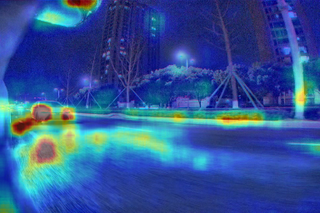}}%
    {\includegraphics[height=1.17cm]{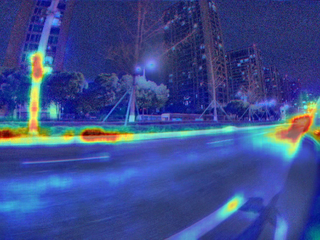}}%
    {\includegraphics[height=1.17cm]{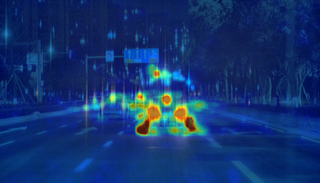}}%
    {\includegraphics[height=1.17cm]{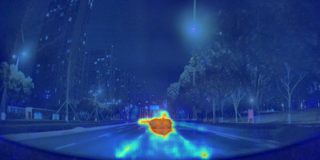}}%
    {\includegraphics[height=1.17cm]{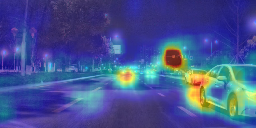}}%
    {\includegraphics[height=1.17cm]{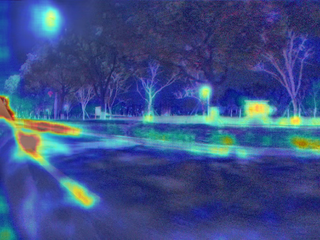}}%
    {\includegraphics[height=1.17cm]{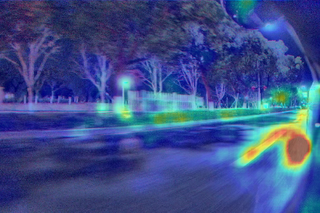}}%
  \vspace{1.2mm}
  
  \figSevenCamRow
    {Camera}%
    {\includegraphics[height=1.17cm]{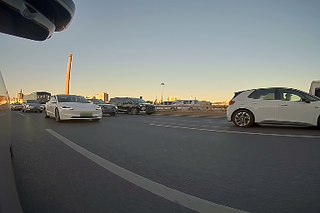}}%
    {\includegraphics[height=1.17cm]{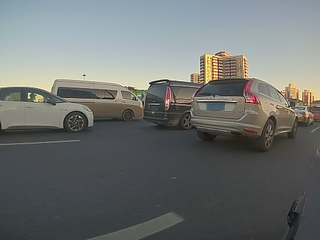}}%
    {\includegraphics[height=1.17cm]{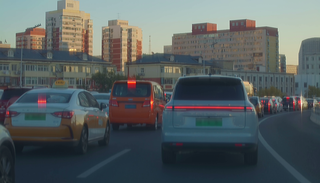}}%
    {\includegraphics[height=1.17cm]{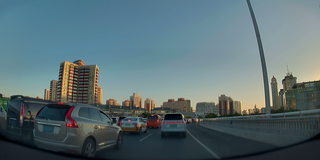}}%
    {\includegraphics[height=1.17cm]{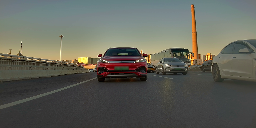}}%
    {\includegraphics[height=1.17cm]{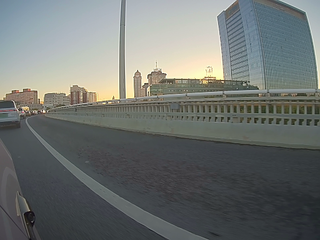}}%
    {\includegraphics[height=1.17cm]{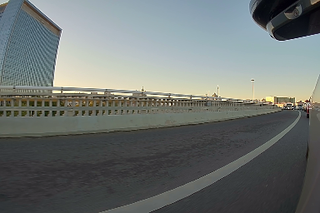}}%
  \figSevenCamRow
    {HybridViT}%
    {\includegraphics[height=1.17cm]{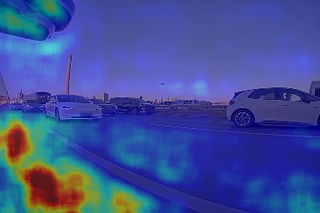}}%
    {\includegraphics[height=1.17cm]{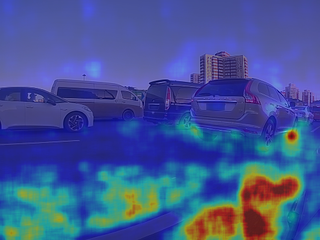}}%
    {\includegraphics[height=1.17cm]{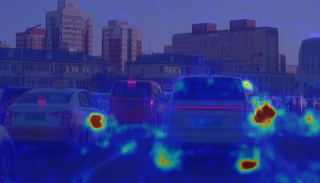}}%
    {\includegraphics[height=1.17cm]{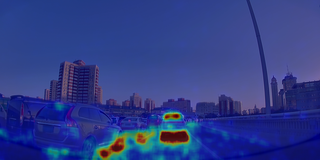}}%
    {\includegraphics[height=1.17cm]{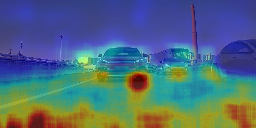}}%
    {\includegraphics[height=1.17cm]{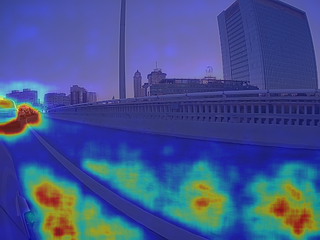}}%
    {\includegraphics[height=1.17cm]{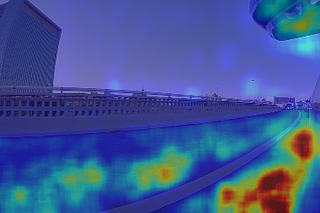}}%
  \figSevenCamRow
    {HybridViT-MLQ}%
    {\includegraphics[height=1.17cm]{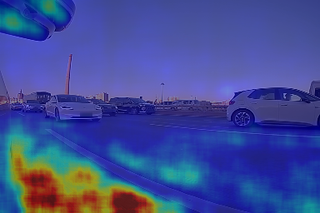}}%
    {\includegraphics[height=1.17cm]{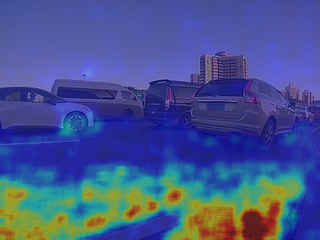}}%
    {\includegraphics[height=1.17cm]{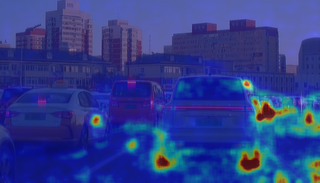}}%
    {\includegraphics[height=1.17cm]{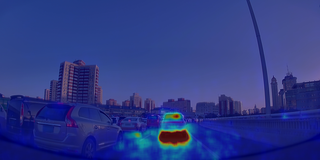}}%
    {\includegraphics[height=1.17cm]{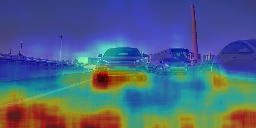}}%
    {\includegraphics[height=1.17cm]{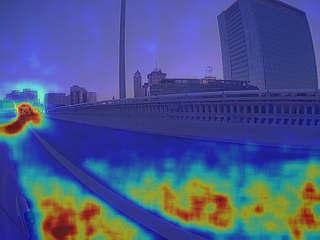}}%
    {\includegraphics[height=1.17cm]{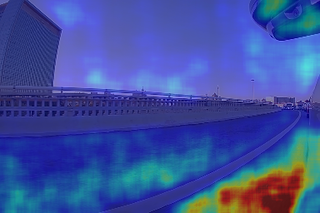}}%
  \caption{\textbf{Integrated Gradients maps for the planned trajectory across seven
  camera views in five driving scenes (top to bottom: split merge, parked-vehicle
  bypass, cut-in, car following, and dense urban-bridge traffic).} Rows show input
  images, HybridViT attributions, and HybridViT-MLQ attributions for the final
  frame of each four-frame input sequence.}
  \label{fig:vla-saliency-appA}
\end{figure}

\end{document}